\documentclass[sigconf]{acmart}

\usepackage{multirow}
\usepackage{colortbl}
\usepackage{tabularx}
\usepackage{array}
\usepackage{float}
\usepackage{placeins}
\usepackage[most]{tcolorbox}

\newtcolorbox{supppromptbox}[1][]{enhanced,breakable,colback=gray!4,
  colframe=gray!55,boxrule=0.45pt,arc=1pt,#1}
\newtcolorbox{qualcasebox}{enhanced,breakable,colback=white,
  colframe=black!55,boxrule=0.5pt,arc=1pt,left=6pt,right=6pt,top=6pt,bottom=6pt}
\newtcolorbox{positivecotbox}{enhanced,breakable,colback=green!3,
  colframe=green!55!black,boxrule=0.55pt,arc=1pt,left=5pt,right=5pt,top=5pt,bottom=5pt}
\newtcolorbox{negativecotbox}{enhanced,breakable,colback=red!3,
  colframe=red!65!black,boxrule=0.55pt,arc=1pt,left=5pt,right=5pt,top=5pt,bottom=5pt}
\newtcolorbox{cotcontentbox}{enhanced,breakable,colback=white,
  colframe=gray!35,boxrule=0.35pt,arc=0.5pt,left=4pt,right=4pt,top=4pt,bottom=4pt}

\renewcommand\footnotetextcopyrightpermission[1]{}

\title{UMER: Unifying Embedding and Ranking with Pair-Aware Discriminative Reasoning for Universal Multimodal Retrieval}
\author{Libiao Chen}
\authornote{These authors contributed equally.}
\affiliation{%
  \institution{Beihang University}
  \city{Beijing}
  \country{China}}
\email{lbchen@buaa.edu.cn}

\author{Xiyang Liu}
\authornotemark[1]
\affiliation{%
  \institution{Beihang University}
  \city{Beijing}
  \country{China}}
\email{xiyangliu@buaa.edu.cn}

\author{Yanheng Wei}
\affiliation{%
  \institution{Independent Researcher}
  \country{China}}

\author{Zhenyu Tang}
\affiliation{%
  \institution{Beihang University}
  \city{Beijing}
  \country{China}}
\email{tangzhenyu@buaa.edu.cn}

\begin{document}

\begin{abstract}
Modern Web information access increasingly involves heterogeneous multimodal content, including images, videos and visually rich documents.
Users may seek not only globally similar content but also candidates that satisfy fine-grained attributes, relations or answer evidence under complex instructions.
Supporting such access requires both efficient corpus-scale retrieval and fine-grained relevance reasoning.
Multimodal large language model (MLLM)-based embedding methods derive representations from MLLM hidden states, and recent work further uses Chain-of-Thought (CoT) reasoning to improve embedding capability by injecting intermediate semantic evidence into the representation. However, existing CoT methods typically use item-wise reasoning over queries and candidates in isolation, providing no explicit evidence to distinguish a positive from a semantically confusable hard negative.
Moreover, contrastive embeddings capture global similarity but struggle with meta-tasks requiring answer verification, category judgment or fine-grained reasoning.
In this paper, we propose \textbf{UMER}, a \textbf{U}nified \textbf{M}ultimodal \textbf{E}mbedding and \textbf{R}anking framework for universal multimodal retrieval.
UMER replaces item-wise reflection with Pair-Aware Discriminative Reasoning, which compares query--candidate pairs to identify instruction-relevant matching and discrepancy evidence.
UMER jointly learns contrastive embeddings for efficient global matching and discriminative ranking for explicit pairwise relevance judgment within a single MLLM.
A complementary mutual distillation strategy further transfers reliable pairwise preferences between the embedding and ranking functions. On MMEB-V2, UMER-R scores 65.5/70.6 with Qwen2-VL-2B/7B, outperforming the strongest included matched-backbone baselines by 1.4/2.0 points. UMER also supports budget-adjustable inference from efficient embedding retrieval to optional pair-aware reranking.

\end{abstract}

\maketitle
\pagestyle{plain}

\section{Introduction}

\begin{figure}[!t]
\centering
\IfFileExists{Figures/intro.png}{\includegraphics[width=\linewidth]{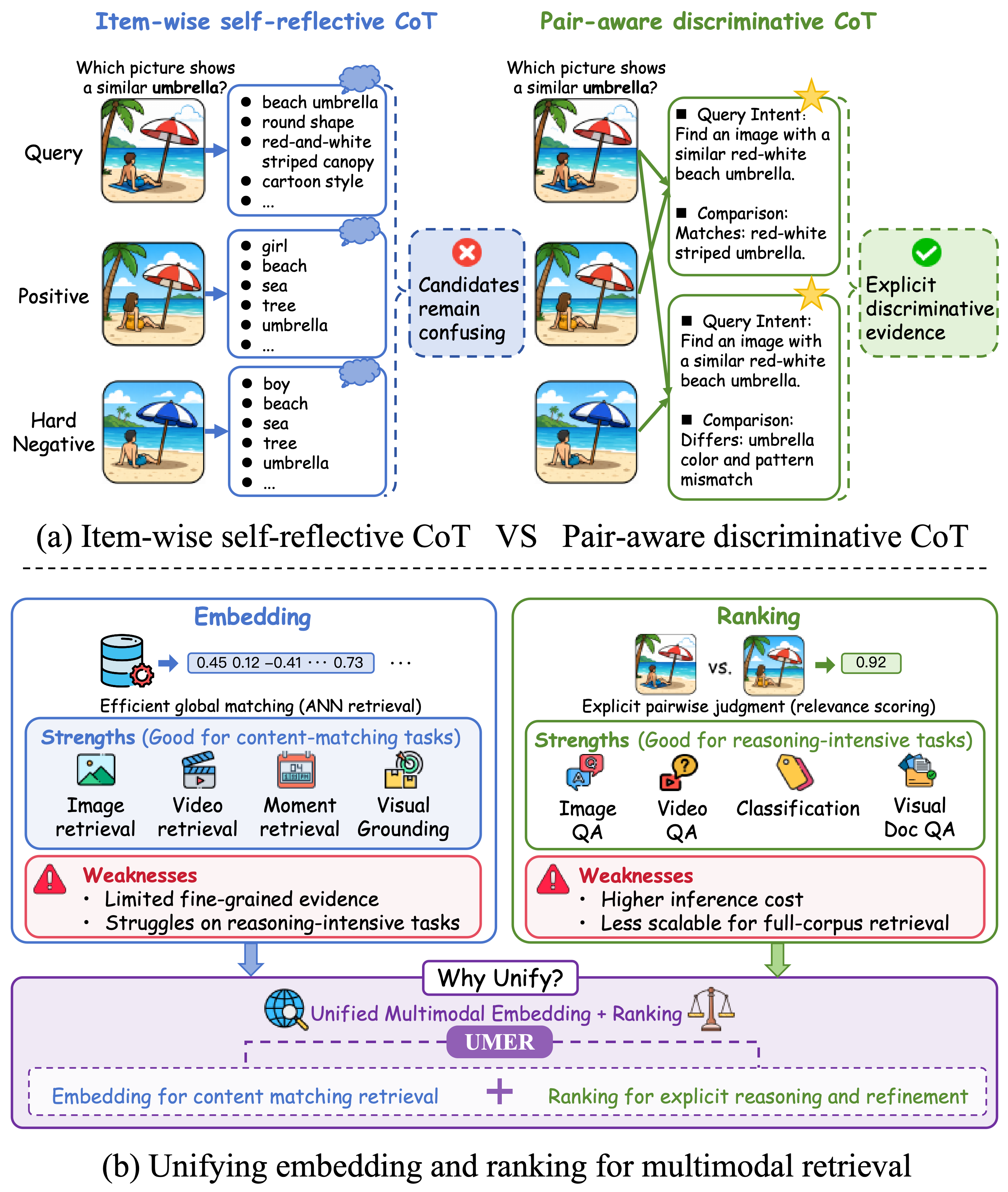}}{} 
\caption{Two overlooked issues in universal multimodal
retrieval. (a) Item-wise self-reflective CoT lacks pair-aware
evidence for hard-negative discrimination. (b) Different meta-
tasks require different capabilities, with embedding and
ranking offering complementary strengths.}
\label{figure_intro}
\end{figure}

Modern Web information access increasingly operates over heterogeneous multimodal content, including images, videos and visually rich documents.
Under complex user instructions, a relevant result may need to satisfy fine-grained attributes, relations, answer evidence or task-specific conditions rather than merely exhibit global semantic similarity.
Supporting such information access over large corpora therefore requires two complementary capabilities: independently indexable representations for efficient retrieval and interaction-based reasoning for precise relevance verification.
Universal multimodal retrieval provides a unified formulation of this challenge across content types and tasks.
Early multimodal representation models, such as CLIP \citep{radford2021clip}, ALIGN \citep{jia2021align} and SigLIP \citep{zhai2023siglip}, typically use dual encoders to align visual and textual modalities and excel at conventional cross-modal content retrieval.
Beyond dual encoders, ALBEF and BLIP combine contrastive alignment with cross-modal fusion or generative objectives \citep{li2021albef,li2022blip}, while ImageBind extends shared embedding spaces beyond image and text to six modalities \citep{girdhar2023imagebind}.
Recently, M-BEIR \citep{wei2024uniir}, MMEB and MMEB-V2 \citep{jiang2024vlm2vec, meng2025vlm2vecv2} have introduced more challenging unified retrieval benchmarks.
These benchmarks cover a broad range of meta-tasks, including classification, question answering, visual grounding, image-level and video-level retrieval and visual document retrieval, imposing higher requirements on content matching, reasoning-intensive relevance judgment and cross-task generalization.
Recent studies apply multimodal large language models (MLLMs) \citep{wang2024qwen2vl} to universal multimodal embedding for complex instruction-aware retrieval.
Most MLLM-based embedding methods \citep{jiang2024e5v,zhang2025gme,lan2025llave,gu2025unime} extract a global representation from a special-token hidden state and optimize it with contrastive learning.
To exploit LLM semantic understanding and reasoning, some works \citep{cui2025tte,wang2026mmembr1} introduce Chain-of-Thought (CoT) \citep{wei2022chain} into multimodal retrieval.
These methods generate textual rationales and derive embeddings from the resulting hidden states, allowing representations to incorporate both the original multimodal inputs and intermediate reasoning.

However, by examining how CoT is currently used in multimodal retrieval and what capabilities complex instruction-aware retrieval tasks require, we identify two overlooked issues, as illustrated in Fig.~\ref{figure_intro}.
\textbf{First, existing item-wise CoT lacks pair-aware discriminative evidence for representation learning.}
Existing CoT-based embedding methods \citep{lan2026umer1,jiang2026embedrl,he2026plume} usually adopt an item-wise self-reflective reasoning paradigm, where the model independently describes the query or candidate.
Although such CoT enriches item-level semantics, it does not model query--candidate interactions and cannot explain why a positive matches or a hard negative fails.
This is especially problematic for hard samples, which often share similar visual content, textual semantics or scene structures, but differ only in subtle attributes, relations, answer evidence or task intent.
Moreover, CoT is typically supervised by autoregressive cross-entropy and only indirectly injected into the embedding token, causing a mismatch between language generation quality and embedding discriminability. 
\textbf{Second, different meta-tasks in universal multimodal retrieval require different model capabilities.}
Specifically, these meta-tasks range from content-matching tasks, such as image retrieval, video retrieval, moment retrieval and visual grounding, which primarily rely on global query--candidate similarity, to reasoning-intensive tasks, such as question answering and classification, which require answer evidence or category-specific conditions.
A single contrastive embedding objective is therefore insufficient to model task-specific matching evidence and fine-grained discrepancies.
In contrast, interaction-based discriminative ranking \citep{lin2025mmembed,gu2026unimev2,li2026qwen3vl} explicitly models pairwise interactions and task satisfaction, complementing efficient embedding retrieval with stronger semantic judgment.
As further corroborated by Fig.~\ref{fig:cmd_transfer}(a), the task-dependent complementarity between embedding and ranking suggests that universal multimodal retrieval should combine representation-based retrieval with explicit discriminative ranking.

We propose \textbf{UMER}, a unified framework for universal multimodal embedding and ranking.
Pair-Aware Discriminative CoT processes positive and hard-negative query--candidate pairs separately to extract instruction-specific evidence. A shared MLLM jointly learns independently indexable embeddings and pair-aware relevance judgments. Complementary Mutual Distillation (CMD) transfers selected preferences between the two functions.
At inference, \textsc{UMER-E} performs corpus-scale vector retrieval without autoregressive reasoning; \textsc{UMER-R} additionally generates pair-aware CoT and relevance scores for the retrieved candidates. These two settings support different accuracy--efficiency requirements.
Their shared backbone allows \textsc{UMER-R} to reuse retrieval-stage query-prefix KV states, avoiding repeated prefill for long multimodal queries. The main contributions are summarized as follows:

\begin{itemize}
    \item We propose Pair-Aware Discriminative Reasoning and construct pair-aware CoT and ranking data, enabling explicit modeling of matching evidence for positives and discrepancy evidence for hard negatives.
    \item We present UMER, a unified MLLM-based framework that jointly learns independently indexable embeddings and pair-aware ranking through a shared backbone with branch-wise attention masks.
    \item We introduce Complementary Mutual Distillation between embedding and ranking: discriminative ranking refines the embedding space, while global embedding structure stabilizes ranking.
    \item Against matched Qwen2-VL baselines, UMER-R obtains the highest aggregate MMEB-V2 scores, while UMER-E retains efficient independently indexable retrieval.
\end{itemize}

\begin{figure*}[!t]
\centering
\IfFileExists{Figures/framework.pdf}{\includegraphics[width=1.0\linewidth]{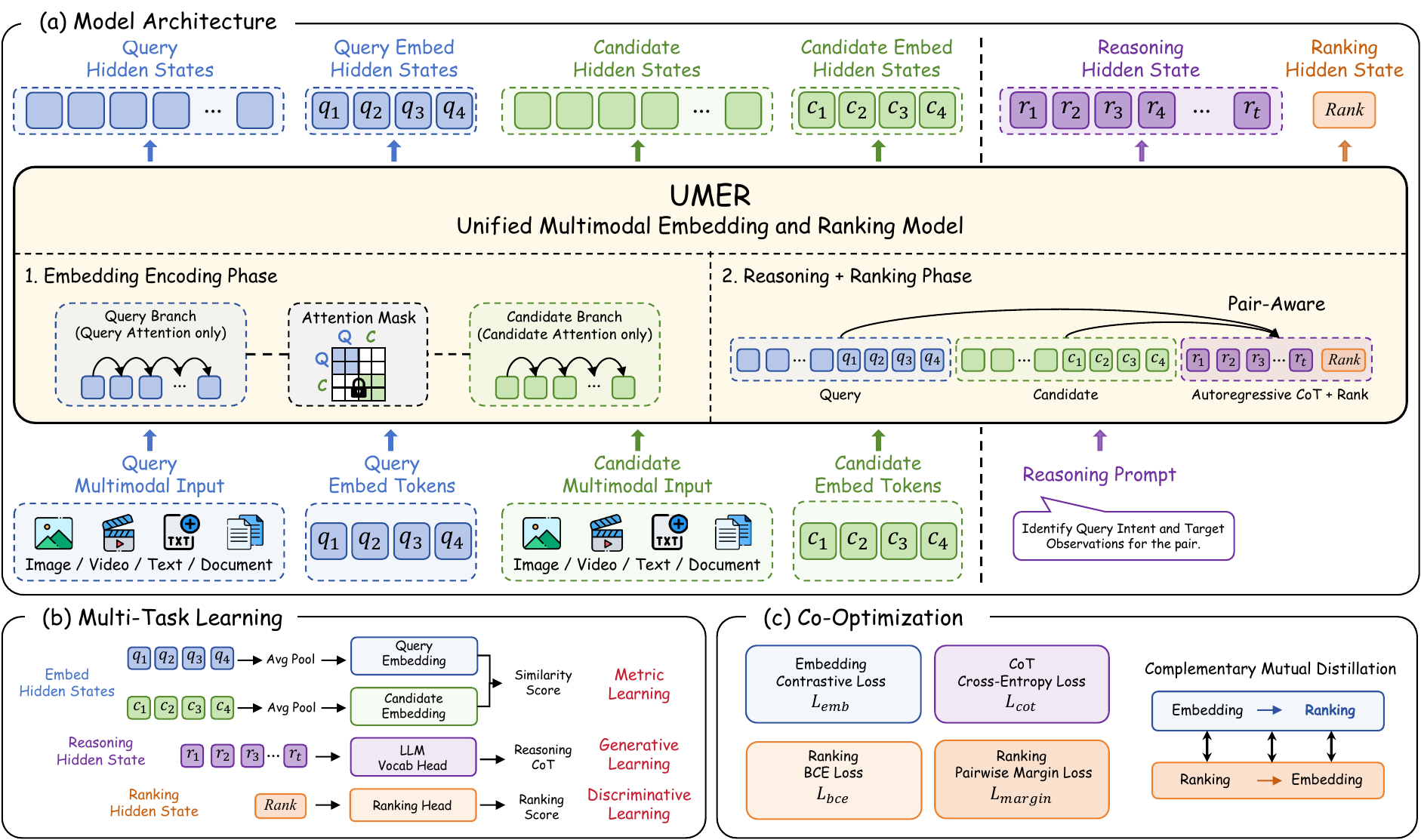}}{} 
\caption{UMER overview.
(a) Masked query/candidate encoding supports independent indexing, followed by pair-aware reasoning and ranking.
(b) Embedding, CoT and relevance objectives train a shared backbone.
(c) Complementary mutual distillation transfers preferences between embedding and ranking.
}
\label{figure_framework}
\end{figure*}

\section{Related Work}

\subsection{Multimodal Embedding Models}
Early multimodal retrieval learns aligned representation spaces through dual encoders such as CLIP, ALIGN and SigLIP \citep{radford2021clip,jia2021align,zhai2023siglip}. ALBEF and BLIP add cross-modal fusion or generative supervision \citep{li2021albef,li2022blip}, while ImageBind extends shared representations beyond image and text \citep{girdhar2023imagebind}.
Instruction-guided universal retrieval is introduced by UniIR and M-BEIR, whose eight task families span diverse query and target modalities beyond symmetric image--text matching \citep{wei2024uniir}. VLM2Vec and VLM2Vec-V2 subsequently adapt MLLMs into general-purpose embedders through contrastive training on MMEB and MMEB-V2 \citep{jiang2024vlm2vec,meng2025vlm2vecv2}. UniME-V2 improves hard-negative supervision through MLLM-based mining and soft-label alignment \citep{gu2026unimev2}, while Qwen3-VL-Embedding combines large-scale contrastive training with reranker distillation \citep{li2026qwen3vl}.
Their shared strength is efficient maximum-inner-product search over precomputed candidate embeddings, which is indispensable for Web-scale retrieval. However, their metric objectives provide limited pair-specific evidence for distinguishing semantically confusable candidates.

\subsection{Reasoning-Enhanced Multimodal Retrieval}
Reasoning-enhanced embedding methods generate intermediate rationales before extracting representations.
UME-R1 derives embeddings from hidden states produced after autoregressive reasoning \citep{lan2026umer1}, and Embed-RL aligns such traces with retrieval rewards through reinforcement learning \citep{jiang2026embedrl}. Because verbalized rationales add sequential decoding before embedding extraction, PLUME reduces them to short continuous-state rollouts \citep{he2026plume}, while LaME uses learnable reason tokens under an information bottleneck \citep{wu2026lame}.
Although these methods differ in reasoning form and cost, they remain item-wise: queries and candidates are reasoned over independently. Consequently, a generated trace never observes the competing candidate that determines the ranking decision, and cannot directly identify the evidence that supports a match or separates a hard negative.

\subsection{Multimodal Reranking Models}
Retrieve-then-rerank pipelines combine independently indexable encoders with interaction-based relevance models. Because interaction scores require joint query--candidate encoding, rerankers operate on a limited candidate set after initial retrieval. This design is established in text retrieval \citep{karpukhin2020dpr,qu2021rocketqa}, including joint retriever--reranker optimization through listwise distillation in RocketQAv2 \citep{ren2021rocketqav2}. In multimodal retrieval, RagVL instruction-tunes MLLMs for knowledge-enhanced image filtering in retrieval-augmented generation \citep{chen2024mllmreranker}, whereas LamRA trains separate LoRA adapters for retrieval and reranking rather than a jointly optimized backbone \citep{liu2024lamra}.
UMER instead uses one masked backbone for independently indexable embeddings and pair-aware ranking, with rationale supervision and selective preference transfer connecting the two functions.



\section{Methodology}

\textbf{Method overview.}
Given an instruction-aware multimodal query $q$ and a corpus of multimodal candidates $\mathcal{C}$, universal multimodal retrieval ranks candidates in $\mathcal{C}$ by their relevance to the query instruction.
As illustrated in Fig.~\ref{figure_framework}, UMER jointly learns two complementary functions with a single shared MLLM: an independently computable embedding function for corpus-scale candidate retrieval and a pair-aware ranking function for explicit verification.
The two functions share the MLLM backbone while using attention masks and task-specific tokens to preserve distinct information flows.
The following subsections present the unified architecture, pair-aware CoT data construction, complementary mutual distillation and training and inference procedures.

\subsection{Unified Embedding and Ranking}

\textbf{Embedding encoding phase.}
Let $Q$ and $C$ denote the tokenized multimodal query and candidate, respectively.
We denote a relevant candidate by $c^+$ and a semantically confusable but irrelevant candidate by $c^-$.
We append $M$ learnable embedding tokens to each input, $E_q=\{q_1,\ldots,q_M\}$ for the query and $E_c=\{c_1,\ldots,c_M\}$ for the candidate, following recent MLLM retrievers that use dedicated tokens as explicit aggregation interfaces \citep{sun2026btoks}.
As shown in Fig.~\ref{figure_framework}(a), under branch-wise attention masks, $E_q$ attends only to $Q$ and $E_c$ only to $C$.
This enables packed training while preserving the independent query and candidate encodings required for corpus indexing and approximate nearest-neighbor search.
We mean-pool and $\ell_2$-normalize the $M$ embedding-token states:
\begin{equation}
e_x=\mathrm{Norm}\!\left(\frac{1}{M}\sum_{k=1}^{M}h_{x_k}\right),
\qquad x\in\{q,c\}.
\label{eq:embeddings}
\end{equation}
where $h_{x_k}$ denotes the final hidden state of the $k$-th embedding token for $x=q$ (query) or $x=c$ (candidate).
The $M$ tokens provide multiple aggregation slots whose pooled state forms a single fixed-dimensional embedding.
We optimize the independently computable representations using an in-batch contrastive objective augmented with mined hard negatives, following established dense-retrieval practice \citep{qu2021rocketqa}:
\begin{equation}
\mathcal{L}_{\rm emb}=-\frac{1}{B}\sum_{i=1}^{B}
\log\frac{\exp(s^e(q_i,c_i^+))}
{\sum_{c\in\{c_i^+\}\cup\mathcal{N}_i}\exp(s^e(q_i,c))},
\label{eq:embedding_loss}
\end{equation}
where $s^e(q,c)=e_q^\top e_c/\tau_e$ is the temperature-scaled cosine similarity, $\tau_e$ is a temperature hyperparameter, $B$ is the batch size and $\mathcal{N}_i$ contains in-batch candidates and mined hard negatives for $q_i$.
This objective learns a shared metric space that supports independent corpus indexing,
while the attention masks ensure that each embedding remains independent of cross-item information unavailable at retrieval time.

\textbf{Pair-aware reasoning and ranking phase.}
Global similarity may fail to distinguish candidates that share visual content or surface semantics but differ in instruction-critical evidence. We therefore introduce pair-aware discriminative reasoning for fine-grained relevance estimation.
For each positive or hard-negative pair $(q,c^a)$, where $a\in\{+,-\}$, we first encode its inputs under branch-wise attention masks, then append a fixed prompt $P$ and autoregressively generate a pair-aware reasoning trace $R^a=(r^a_1,\ldots,r^a_{T_a})$ with full pairwise attention.
Let $X^a=[Q;E_q;C^a;E_c;P]$ denote the resulting input sequence, where $C^a$ is the tokenized candidate, and let $g^a$ be the verified target reasoning trace.
We supervise the generation of reasoning traces with
\begin{equation}
\mathcal{L}_{\rm cot}=\sum_{a\in\{+,-\}}
\mathrm{CE}\!\left(g^a;X^a\right),
\label{eq:cot_loss}
\end{equation}
where $\mathrm{CE}(g^a;X^a)$ is the standard token-level cross-entropy under the model's autoregressive distribution. 
This objective trains pair-aware discriminative CoT to ground its reasoning in jointly visible query--candidate evidence, rather than independently describing each item as in item-wise self-reflective CoT.
During training, we teacher-force the target rationale; at inference, we decode it autoregressively.

The rationale-conditioned pair representation is then converted into a relevance judgment by a learnable $\mathtt{[RANK]}$ token following $R^a$, which attends to the full pair and produces a logit $z_{q,c^a}=w^{\top}h_{\mathtt{[RANK]}}+b$, with probability $p_{q,c^a}=\sigma(z_{q,c^a})$.
We train the positive and hard-negative pairs with binary cross-entropy,
\begin{equation}
\mathcal{L}_{\rm bce}=-\log p_{q,c^+}-\log\left(1-p_{q,c^-}\right),
\label{eq:bce_loss}
\end{equation}
and use a logistic pairwise objective to rank the positive above the hard negative:
\begin{equation}
\mathcal{L}_{\rm margin}=-\log\sigma\!\left(z_{q,c^+}-z_{q,c^-}\right).
\label{eq:rank_loss}
\end{equation}
We average each loss over valid triplets $(q,c^+,c^-)$ in the minibatch.
This smooth objective encourages $z_{q,c^+}>z_{q,c^-}$.
$\mathcal{L}_{\rm cot}$ supervises the pair-conditioned discriminative evidence, 
$\mathcal{L}_{\rm bce}$ provides absolute relevance labels,
and $\mathcal{L}_{\rm margin}$ encourages the positive logit to exceed the hard-negative logit.

\subsection{Pair-Aware Discriminative CoT Data Construction}

Item-wise CoT \citep{lan2026umer1,jiang2026embedrl,he2026plume} can richly describe individual items while still omitting the evidence needed to distinguish a positive from a semantically confusable negative.
We therefore construct pair-aware CoT and ranking supervision through three steps: hard-negative mining, structured CoT generation and evidence-sufficiency filtering.


\textbf{Hard-negative mining.}
Given the training set $\mathcal{D}=\{(q_i,c_i^+)\}_{i=1}^{N}$, we use a frozen off-the-shelf multimodal embedder $f_0$ to obtain normalized representations for every query and candidate.
For each query $q_i$, we rank non-relevant candidates $\mathcal{C}_i^-$ by cosine similarity and retain the three highest-scoring candidates:
\begin{equation}
\mathcal{H}_i=\mathrm{Top3}_{c\in\mathcal{C}_i^-}\,
f_0(q_i)^{\top}f_0(c).
\label{eq:hard_negative_mining}
\end{equation}
Each query thus yields one positive pair $(q_i,c_i^+)$ and three hard-negative pairs $(q_i,c^-_{i,j})$, which serve as candidates for verified triplet construction.

\textbf{Structured CoT generation.}
For each positive or mined hard-negative pair, we prompt Qwen3.5-9B \citep{qwen2026qwen35} to generate structured pair-aware CoT. The CoT contains two fields: \textbf{Query Intent} and \textbf{Target Observations}.
Query Intent specifies the evidence required by the instruction, whereas Target Observations identifies candidate evidence that supports or contradicts that intent.
This structure instantiates the pair-aware reasoning paradigm in Fig.~\ref{figure_intro}(a) by evaluating a shared query intent against each positive or hard-negative candidate.
We omit an explicit Yes/No answer so that relevance must be inferred from the CoT evidence rather than copied from a label.
The text-only check tests whether the generated evidence supports the annotated relevance label; it does not independently verify visual observations against the original input.


\textbf{Evidence-sufficiency filtering.}
A fluent CoT does not necessarily contain sufficient discriminative evidence.
We therefore retain a generated CoT only if a lightweight text-only verifier (Qwen3.5-0.8B) can correctly infer the pair's ground-truth Yes/No relevance label from the CoT alone. For each mined hard negative $c^-_{i,j}$, we retain the triplet $(q_i,c_i^+,c^-_{i,j})$ only if the CoTs for both the positive pair and its corresponding hard-negative pair pass this verification.
Applying this generate-and-verify pipeline to all training queries yields verified triplets for CoT generation, discriminative ranking and distillation.
More details are provided in the appendix.


\subsection{Complementary Mutual Distillation}

Joint retriever--reranker training can transfer relevance information between two retrieval stages \citep{ren2021rocketqav2}. Embedding and ranking nevertheless capture complementary relevance signals in our setting: embedding organizes global semantic similarity, whereas pair-aware ranking resolves instruction-critical confusions. Rather than forcing the two functions to agree indiscriminately, we use selective bidirectional CMD to transfer a preference only to the \emph{other} capability and only on meta-tasks where its source is more reliable. This preserves their complementary roles while allowing each to address the other's blind spots.

For each verified triplet, a frozen reference model produces temperature-scaled positive--negative preference distributions $\bar{\pi}_e$ and $\bar{\pi}_r$ for embedding and ranking, respectively; $\pi_e$ and $\pi_r$ are the corresponding distributions from the trainable model.  We assign each meta-task to an embedding-favored or ranking-favored set according to its task attribute: Retrieval and Grounding/MR are treated as content-matching tasks where global embedding structure is the primary teacher, while Classification and QA are treated as reasoning-intensive tasks where pair-aware ranking provides the primary teacher.  The gate $g_{e\rightarrow r}$ is one only when an embedding-favored triplet is correctly ordered by the reference embedding; analogously, $g_{r\rightarrow e}$ is one only when a ranking-favored triplet is correctly ordered by the reference ranker.  Both gates are zero otherwise, preventing unreliable preferences from propagating.

\begin{equation}
\begin{array}{rcl}
\mathcal{L}_{\rm cmd}&=&g_{e\rightarrow r}\,\mathrm{KL}\!\left(\mathrm{sg}[\bar{\pi}_e]\|\pi_r\right)\\
&&+g_{r\rightarrow e}\,\mathrm{KL}\!\left(\mathrm{sg}[\bar{\pi}_r]\|\pi_e\right),
\end{array}
\label{eq:cmd_loss}
\end{equation}
where $\mathrm{sg}[\cdot]$ stops gradients through the frozen reference, and the loss is averaged over activated triplets.  The first term is exclusively embedding-to-ranking distillation, and the second is exclusively ranking-to-embedding distillation.
Consequently, reliable embedding preferences stabilize ranking on content-matching meta-tasks, while reliable ranking preferences refine the embedding space on reasoning-intensive meta-tasks.

\subsection{Progressive Training and Flexible Inference}

\textbf{Three-stage training.}
We train UMER progressively to prevent early generative and ranking objectives from destabilizing the shared backbone, following staged reasoning-to-embedding optimization \citep{he2026plume}.
The overall objective is
\begin{equation}
\begin{array}{rcl}
\mathcal{L}_{\mathrm{all}} &=&\mathcal{L}_{\rm emb}+\lambda_{\rm cot}\mathcal{L}_{\rm cot}
+\lambda_{\rm bce}\mathcal{L}_{\rm bce}\\
&&+\lambda_{\rm margin}\mathcal{L}_{\rm margin}
+\lambda_{\rm cmd}\mathcal{L}_{\rm cmd},
\end{array}
\label{eq:total_loss}
\end{equation}
where the $\lambda$ terms weight the CoT, ranking and CMD losses relative to $\mathcal{L}_{\rm emb}$.
Stage~1 learns a stable, indexable embedding space using only $\mathcal{L}_{\rm emb}$ on the full MMEB training set.
Stage~2 adds $\mathcal{L}_{\rm cot}$, $\mathcal{L}_{\rm bce}$ and $\mathcal{L}_{\rm margin}$ on verified pair-aware triplets to learn discriminative evidence and relevance ordering.
Stage~3 freezes the Stage~2 checkpoint as the reference model and adds $\mathcal{L}_{\rm cmd}$, transferring reliable preferences across capabilities on complementary meta-tasks.

\begin{table*}[t]
\centering
\scriptsize
\setlength{\tabcolsep}{3.0pt}
\renewcommand{\arraystretch}{1.08}
\begin{tabular*}{\textwidth}{@{\extracolsep{\fill}}l*{16}{c}@{}}
\toprule
\multirow[c]{2}{*}{\textbf{Model}} & \multicolumn{5}{c}{\textbf{Image}} & \multicolumn{5}{c}{\textbf{Video}} & \multicolumn{5}{c}{\textbf{VisDoc}} & \textbf{All} \\
\cmidrule(lr){2-6}\cmidrule(lr){7-11}\cmidrule(lr){12-16}
& \textbf{CLS} & \textbf{QA} & \textbf{RET} & \textbf{GD} & \textbf{Overall}
& \textbf{CLS} & \textbf{QA} & \textbf{RET} & \textbf{MRET} & \textbf{Overall}
& \textbf{VDRv1} & \textbf{VDRv2} & \textbf{VR} & \textbf{OOD} & \textbf{Overall} & \\
\midrule
\textbf{\# of Datasets} & 10 & 10 & 12 & 4 & 36 & 5 & 5 & 5 & 3 & 18 & 10 & 4 & 6 & 4 & 24 & 78 \\
\midrule
\rowcolor{black!12}\multicolumn{17}{c}{\textbf{Qwen2-VL-2B Backbones}} \\
\midrule
\rowcolor{black!7}\multicolumn{17}{c}{\textit{Baseline Models}} \\
GME
& 54.4 & 29.9 & 66.9 & 55.5 & 51.9
& 34.9 & 42.0 & 25.6 & 32.4 & 33.9
& \textbf{86.1} & \textbf{54.0} & 82.5& 43.1 & \underline{72.7} & 54.1 \\
VLM2Vec
& 58.7 & 49.3 & 65.0 & 72.9 & 59.7
& 33.4 & 30.5 & 20.6 & 33.0 & 29.0
& 49.8 & 13.5 & 51.8 & 33.5 & 41.6 & 47.0 \\
VLM2Vec-V2
& 62.9 & 56.3 & 69.5 & 77.3 & 64.9
& 39.3 & 34.3 & 28.8 & 38.5 & 34.9
& 75.5 & 44.9 & 79.4 & 39.4 & 65.4 & 58.0 \\
DUME
& 59.3 & 55.0 & 66.3 & 78.0 & 62.5
& 37.7 & 46.6 & 17.1 & 30.0 & 33.2
& 67.6 & 43.3 & 47.1 & 33.8 & 52.8 & 52.7 \\
BToks
& 64.3 & 59.8 & 68.8 & 77.4 & 66.0
& 43.7 & 47.0 & 33.0 & 33.6 & 39.9
& 71.1 & 38.6 & 81.3 & 38.1 & 62.7 & 59.0 \\
UME-R1
& 64.8 & 62.8 & 67.6 & 77.2 & 66.6
& 44.3 & 51.2 & 32.9 & \underline{39.7} & 42.2
& 72.4 & 46.2 & 79.2 & 37.2 & 63.9 & 60.1 \\
PLUME
& \underline{66.5} & 59.2 & 67.6 & 79.7 & 66.3
& 45.0 & \underline{52.3} & 33.5 & \textbf{46.7} & \underline{44.1}
& 72.1 & 49.8 & 78.1 & 57.4 & 67.5 & 61.6 \\
RIME
& \textbf{67.9} & \underline{64.4} & 69.8 & \textbf{82.1} & \underline{69.1}
& \underline{48.0} & 52.1 & 33.6 & 39.2 & 43.7
& 76.4 & \underline{51.4} & 81.7 & 63.9 & 71.4 & \underline{64.1} \\
\midrule
\rowcolor{black!7}\multicolumn{17}{c}{\textit{Ours}} \\
\textsc{UMER-E}     & 64.4 & 64.9 & \underline{70.2} & 78.4 & 68.0 & 43.0 & 50.6 & \underline{34.7} & 32.8 & 41.1 & 76.1 & 50.1 & \underline{82.8} & \underline{68.3} & 72.2 & 63.1 \\
\textsc{UMER-R}     & 66.1 & \textbf{68.9} & \textbf{71.7} & \underline{80.8} & \textbf{70.4} & \textbf{50.6} & \textbf{52.8} & \textbf{37.7} & 35.0 & \textbf{45.0} & \underline{78.0} & 50.5 & \textbf{85.0} & \textbf{68.8} & \textbf{73.6} & \textbf{65.5} \\
\midrule
\rowcolor{black!12}\multicolumn{17}{c}{\textbf{Qwen2-VL-7B Backbones}} \\
\midrule
\rowcolor{black!7}\multicolumn{17}{c}{\textit{Baseline Models}} \\
GME
& 57.7 & 34.7 & 71.2 & 59.3 & 56.0
& 37.4 & 50.4 & 28.4 & 37.0 & 38.4
& \textbf{89.4} & 55.6 & 85.0 & 44.4 & 75.2 & 57.8 \\
VLM2Vec
& 62.7 & 56.9 & 69.4 & 82.2 & 65.5
& 39.1 & 30.0 & 29.0 & 38.9 & 33.7
& 56.9 & 9.4 & 59.1 & 38.1 & 46.4 & 52.3 \\
DUME
& 64.2 & 57.0 & 70.8 & 81.8 & 66.4
& 32.9 & 47.4 & 8.6 & 28.0 & 29.4
& 67.1 & 35.2 & 82.6 & 34.9 & 60.3 & 55.9 \\
UME-R1
& 67.1 & 69.2 & 71.9 & 84.9 & 71.3
& 48.6 & \underline{60.7} & 38.2 & 39.3 & 47.5
& 75.7 & 50.5 & 83.7 & 37.6 & 67.1 & 64.5 \\
RIME
& \textbf{70.3} & \underline{71.7} & \underline{73.2} & \underline{86.3} & \underline{73.4}
& \underline{52.6} & \textbf{62.0} & \underline{38.4} & \underline{41.6} & \underline{49.4}
& 80.9 & 55.6 & \underline{85.8} & 66.9 & 75.6 & \underline{68.6} \\
\midrule
\rowcolor{black!7}\multicolumn{17}{c}{\textit{Ours}} \\
\textsc{UMER-E}     & 66.2 & 69.7 & 72.3 & 84.6 & 71.2 & 48.0 & 54.7 & 35.6 & 40.7 & 45.2 & 81.3 & \underline{57.8} & 85.1 & \underline{72.5} & \underline{76.8} & 67.0 \\
\textsc{UMER-R}     & \underline{68.0} & \textbf{73.7} & \textbf{75.6} & \textbf{87.7} & \textbf{74.3} & \textbf{55.2} & \underline{60.7} & \textbf{40.5} & \textbf{42.5} & \textbf{50.5} & \underline{85.0} & \textbf{62.1} & \textbf{87.9} & \textbf{73.8} & \textbf{80.0} & \textbf{70.6} \\
\bottomrule
\end{tabular*}
\caption{Main results on MMEB-V2, grouped by Qwen2-VL backbone size. Within each size block, the best and second-best scores in each column are in \textbf{bold} and \underline{underlined}, respectively. Baseline values are reported results from the cited papers, whereas UMER values are computed over all 78 tasks using the corrected benchmark.}
\label{tab:main_results}
\end{table*}

\textbf{Flexible inference.}
UMER supports two budget-adjustable inference settings:
\textsc{UMER-E} for embedding-only retrieval and \textsc{UMER-R} for pair-aware reranking with score combination.
We precompute candidate embeddings with Eq.~\ref{eq:embeddings}, build an approximate nearest-neighbor index and independently encode each query to retrieve the top-$K$ candidates by $s^e(q,c)$;
\textsc{UMER-E} uses this score directly without autoregressive reasoning.
For \textsc{UMER-R}, we apply the pair-aware prompt only to the top-$K$ candidates to obtain the ranking logit $z_{q,c}$, following the standard retrieve-then-rerank paradigm \citep{chen2024mllmreranker,li2026qwen3vl}, and combine it with the first-stage embedding score:
\begin{equation}
s^{\rm r}(q,c)=\widehat{s^e(q,c)}+\alpha\,\widehat{z_{q,c}},
\label{eq:ranking_score}
\end{equation}
where $\widehat{\cdot}$ denotes normalization within the top-$K$ list and $\alpha$ controls the ranking contribution.

\section{Experiments and Results}

We evaluate UMER from four perspectives: overall effectiveness and scaling, the distinct roles of ranking, CoT and CMD, the mechanisms underlying pair-aware supervision and cross-capability transfer, and the resulting accuracy--efficiency tradeoff.

\subsection{Experimental Setup}

\textbf{Benchmark and metrics.}
We evaluate UMER on MMEB-V2 \citep{meng2025vlm2vecv2}, which comprises 78 tasks across image, video and visual-document domains, covering classification, question answering, retrieval, grounding and moment retrieval.
We use the corrected versions of ViDoSeek-page and MMLongBench-page.
For a fair comparison, our training data are drawn from the same public source datasets as VLM2Vec-V2.
Following the official protocol, we report Hit@1 for image and video tasks and NDCG@5 for visual-document retrieval. Each meta-task score is the macro average over its constituent datasets, and \textsc{All} is the macro average over all 78 tasks.
We report the complete dataset-level results for both backbone sizes in the supplementary material.

\textbf{Baselines.}
For the 2B block, we compare with eight representative unified multimodal retrieval methods: GME \citep{zhang2025gme}, VLM2Vec \citep{jiang2024vlm2vec}, VLM2Vec-V2 \citep{meng2025vlm2vecv2}, DUME \citep{lan2026umer1}, Bottleneck Tokens (BToks) \citep{sun2026btoks}, UME-R1 \citep{lan2026umer1}, PLUME \citep{he2026plume} and RIME \citep{wu2026rime}.
For the 7B block, we include GME, VLM2Vec, DUME, UME-R1 and RIME, for which published 7B task-level results are available. Comparisons are restricted to matched Qwen2-VL backbone sizes and do not constitute a global leaderboard comparison.

\textbf{Implementation details.}
Our primary model is initialized from Qwen2-VL-2B-Instruct, and we additionally train a Qwen2-VL-7B counterpart. Each input receives \(M=4\) learnable embedding tokens.
Relative to the embedding objective, the effective weights for CoT generation, binary ranking, pairwise ranking and bidirectional distillation are $(\lambda_{\rm cot},\lambda_{\rm bce},\lambda_{\rm margin},\lambda_{\rm cmd})=(0.2,0.1,0.2,0.2)$.  At inference time, pair-aware reasoning is applied to the top $K=5$ candidates retrieved by the embedding branch, with generation capped at 256 new tokens.  For \textsc{UMER-R}, the embedding and ranking scores are separately z-score normalized over these five candidates, and we set $\alpha=2.0$ in Eq.~\ref{eq:ranking_score}.
Additional details are provided in the supplementary material.

\subsection{Overall Effectiveness and Scaling}
Table~\ref{tab:main_results} evaluates the two operating points supported by one shared model: independently indexable \textsc{UMER-E} and candidate-level \textsc{UMER-R}.
Their overall scores are 63.1/65.5 at 2B and 67.0/70.6 at 7B, with \textsc{UMER-R} ranking first within both matched Qwen2-VL backbone blocks.
Compared with the strongest same-scale baseline, RIME, \textsc{UMER-R} improves the overall score by 1.4 points at 2B and 2.0 points at 7B, with the larger 7B margin showing that pair-aware refinement remains effective as backbone capacity increases.
Relative to \textsc{UMER-E}, candidate-level refinement improves the aggregate by 2.4 and 3.6 points, respectively, with gains across image (+2.4/+3.1), video (+3.9/+5.3) and visual-document (+1.4/+3.2) tasks.
Its additional online cost is quantified in Table~\ref{tab:baseline_efficiency}.


\subsection{Component Roles and Transfer}

\begin{table*}[t]
\centering
\scriptsize
\begin{tabular*}{\textwidth}{@{\extracolsep{\fill}}lcccccccc@{}}
\toprule
\multirow[c]{2}{*}{\textbf{Configuration}} & \multicolumn{4}{c}{\textbf{Embedding setting (UMER-E)}} & \multicolumn{4}{c}{\textbf{Pair-aware setting (UMER-R)}} \\
\cmidrule(lr){2-5}\cmidrule(lr){6-9}
 & \textbf{Image} & \textbf{Video} & \textbf{VisDoc} & \textbf{All} & \textbf{Image} & \textbf{Video} & \textbf{VisDoc} & \textbf{All} \\
\midrule
Embedding only & 66.5 & 37.8 & 69.2 & 60.7 & -- & -- & -- & -- \\
$+$ pair-aware CoT, w/o BCE/margin & \textbf{68.2} & \textbf{41.1} & 71.2 & \underline{62.9} & 68.6 & 42.2 & 71.3 & 63.3 \\
$+$ ranking losses, w/o CoT & \underline{68.0} & 40.2 & 69.3 & 62.0 & \textbf{70.8} & \underline{45.1} & 71.2 & 65.0 \\
$+$ pair-aware CoT and ranking losses & 67.9 & 39.6 & 71.1 & 62.4 & 70.3 & 44.9 & \underline{73.4} & \underline{65.4} \\
\midrule
Pair-aware model, w/o CMD & 67.9 & 39.6 & 71.1 & 62.4 & 70.3 & 44.9 & \underline{73.4} & \underline{65.4} \\
$+$ ranking $\rightarrow$ embedding only & 67.9 & \underline{40.4} & \underline{71.5} & 62.7 & 70.6 & 44.7 & 72.7 & 65.3 \\
$+$ embedding $\rightarrow$ ranking only & 67.8 & \underline{40.4} & 70.1 & 62.2 & \textbf{70.8} & \underline{45.1} & 72.6 & \textbf{65.5} \\
$+$ always-on bidirectional CMD & \underline{68.0} & 40.2 & 71.1 & 62.6 & \underline{70.7} & \textbf{45.3} & 73.0 & \textbf{65.5} \\
$+$ selective bidirectional CMD (full) & \underline{68.0} & \textbf{41.1} & \textbf{72.2} & \textbf{63.1} & 70.4 & 45.0 & \textbf{73.6} & \textbf{65.5} \\
\bottomrule
\end{tabular*}
\caption{Controlled ablations of pair-aware CoT, ranking supervision and complementary mutual distillation (CMD). The best and second-best scores in each column are in \textbf{bold} and \underline{underlined}, respectively.}
\label{tab:ablation}
\end{table*}

\textbf{A division of labor between ranking and CoT.}
Table~\ref{tab:ablation} shows that the CoT-only variant raises \textsc{UMER-E} from 60.7 to 62.9; it appends a Yes/No decision trained autoregressively and uses its log-odds for \textsc{UMER-R}, with the $\mathtt{[RANK]}$ head disabled.
The resulting combined score improves over \textsc{UMER-E} by 0.4 points, showing that the autoregressive decision supplies a modest pairwise signal without the discriminative head.
This control therefore isolates a generative relevance path rather than attributing the 63.3 score to an untrained $\mathtt{[RANK]}$ head.
Direct ranking supervision accounts for a substantial portion of the candidate-level gain.
In this no-CoT control, we retain the same query--candidate pairs, embedding objective and ranking losses, but place the $\mathtt{[RANK]}$ token directly after the pair prompt without rationale generation or conditioning; evaluation otherwise matches \textsc{UMER-R}.
This variant produces 62.0/65.0 for \textsc{UMER-E}/\textsc{UMER-R}.
Its larger gain indicates that explicitly calibrated candidate ordering accounts for most of the reranking benefit.
Under this matched ranking control, adding Pair-Aware CoT further raises both settings by 0.4 points to 62.4/65.4 while providing inspectable evidence for each candidate decision.

\textbf{Complementary mutual distillation.}
CMD primarily transfers capability between the two functions rather than acting as another uniform score booster.
Relative to the pair-aware model without CMD, selective bidirectional CMD improves \textsc{UMER-E} by 0.7 points (62.4 to 63.1) while retaining the best \textsc{UMER-R} score of 65.5, strengthening the efficient operating point rather than merely tuning the final combined score.
The one-way results expose an asymmetric tradeoff: ranking-to-embedding raises \textsc{UMER-E} to 62.7, whereas embedding-to-ranking reaches 65.5 in \textsc{UMER-R} but lowers \textsc{UMER-E} to 62.2.
Always-on CMD also reaches 65.5 in \textsc{UMER-R} but only 62.6 in \textsc{UMER-E}, so selective gating is required to attain 63.1 without sacrificing reranking accuracy.
Because the combined score masks branch-specific changes, Fig.~\ref{fig:cmd_transfer} further analyzes where this transfer occurs.

\subsection{Mechanistic Analysis}

\begin{figure}[t]
\centering
\IfFileExists{Figures/top1_margin_and_confidence_curve_paper.pdf}{\includegraphics[width=\linewidth]{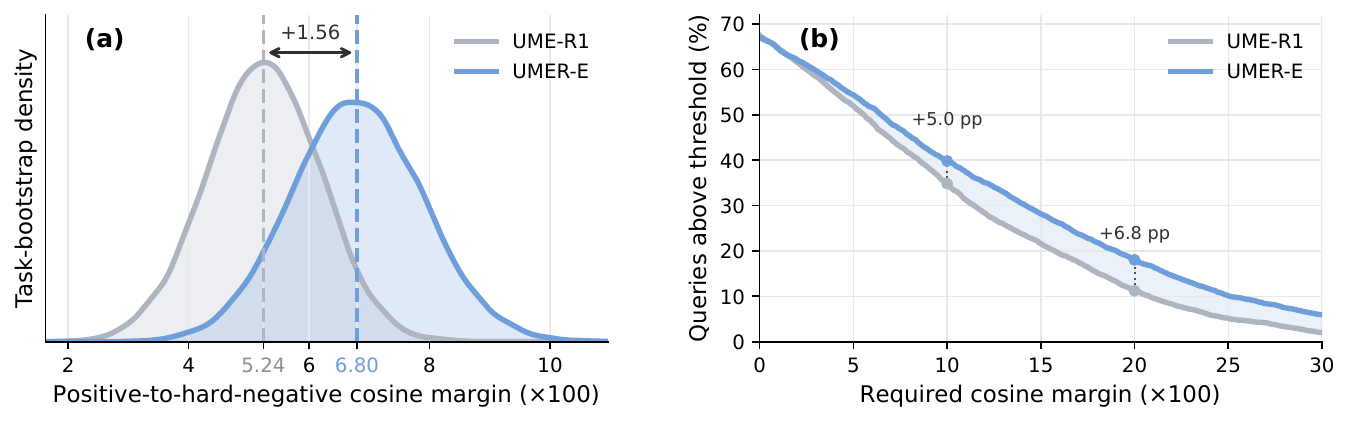}}{}
\caption{
Embedding separation under item-wise and pair-aware supervision.
(a) Bootstrap distribution of the macro-averaged positive--hard-negative margin.
(b) Fraction of queries exceeding margin thresholds.
}
\label{fig:embedding_separation}
\end{figure}

\textbf{Hard-negative separation under pair-aware supervision.}
Both UME-R1 and UMER-E yield independently indexable embeddings, but differ in how these representations are learned: UME-R1 relies on item-wise reasoning, whereas UMER-E is optimized within a pair-aware reasoning and supervision framework.
We compare them on the same 3,600 queries from all 36 image tasks.
For each query, we define the separation margin as the similarity of its highest-scoring positive minus that of its highest-scoring non-positive.
Fig.~\ref{fig:embedding_separation}(a) shows that UMER-E increases the task-macro-averaged positive-to-hard-negative cosine margin from 5.24 to 6.80 points. While the two models achieve comparable positive-margin coverage, Fig.~\ref{fig:embedding_separation}(b) reveals a widening advantage under stricter thresholds: UMER-E retains 5.0 and 6.8 percentage points more queries at margins of 10 and 20, respectively.
These results are consistent with pair-aware supervision producing more decisive separation between successful matches and their hardest distractors.

\begin{figure}[t]
\centering
\IfFileExists{Figures/cmd_stage2_stage3_two_panel_preview.pdf}{\includegraphics[width=\linewidth]{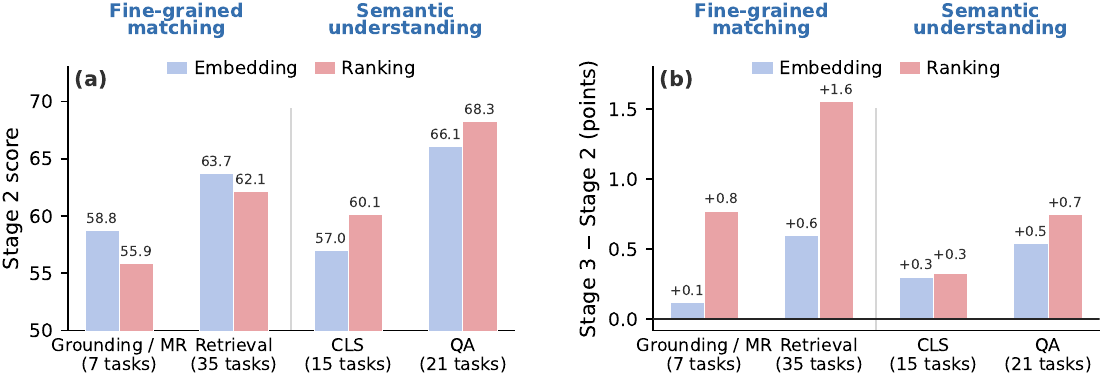}}{}
\caption{Capability specialization and transfer via CMD.
(a) Embedding is stronger for content matching, while ranking is stronger for reasoning-intensive relevance judgment.
(b) CMD transfers these complementary strengths between the two functions.}
\label{fig:cmd_transfer}
\end{figure}

\textbf{From capability specialization to complementary transfer.}
Fig.~\ref{fig:cmd_transfer}(a) reveals a systematic capability split within the same unified Stage~2 model: embedding outperforms ranking on content-matching Retrieval and Grounding/MR, whereas ranking outperforms embedding on reasoning-intensive CLS and QA.
This task-dependent reversal directly supports the motivation in Fig.~\ref{figure_intro}(b): heterogeneous retrieval tasks require distinct inductive biases, and neither metric-space embedding nor pair-aware ranking is uniformly optimal.
Unification is therefore not merely an architectural convenience, but a means to retain and coordinate both capabilities within one model.
Building on this specialization, Fig.~\ref{fig:cmd_transfer}(b) shows directionally aligned transfer: on embedding-favored Retrieval and Grounding/MR, ranking gains 0.8 and 1.6 points, respectively; on ranking-favored CLS and QA, embedding gains 0.3 and 0.5 points.
These targeted branch-level improvements, rather than the modest change in the already combined score alone, provide the main evidence that CMD narrows capability gaps without erasing useful specialization.


\begin{figure*}[t]
\centering
\setlength{\fboxsep}{3.5pt}
\setlength{\fboxrule}{0.45pt}
\newcommand{\cotbullet}[1]{\par\noindent\(\bullet\)~#1}
\newcommand{\casevisualhead}{%
  \parbox[c][10pt][c]{\linewidth}{\centering\footnotesize\bfseries
    \makebox[0.333\linewidth][c]{\textcolor{blue!70!black}{Query}}%
    \makebox[0.333\linewidth][c]{\textcolor{red!70!black}{Top-1 Hard Negative}}%
    \makebox[0.333\linewidth][c]{\textcolor{green!55!black}{Positive Candidate}}}%
  \par\vspace{3pt}}
\newcommand{\casecotevidencehead}{%
  \parbox[c][10pt][c]{\linewidth}{\centering\footnotesize\bfseries
    \textcolor{violet!75!black}{Pair-Aware CoT Evidence}}%
  \par\vspace{3pt}}

\begin{minipage}{\textwidth}
\small
\textbf{Case 1: Fine-grained image composition.}
\textbf{Query:} Find an everyday image of the same breed dog, specifically a Cavalier King Charles Spaniel, focusing on its head.
\par\vspace{3pt}
\begin{minipage}[t]{0.43\textwidth}
\vspace{0pt}
\centering
\casevisualhead
\IfFileExists{supplementary/figures/g-cirr-breed-correction.png}{\includegraphics[width=\linewidth,trim=0 0 0 122,clip]{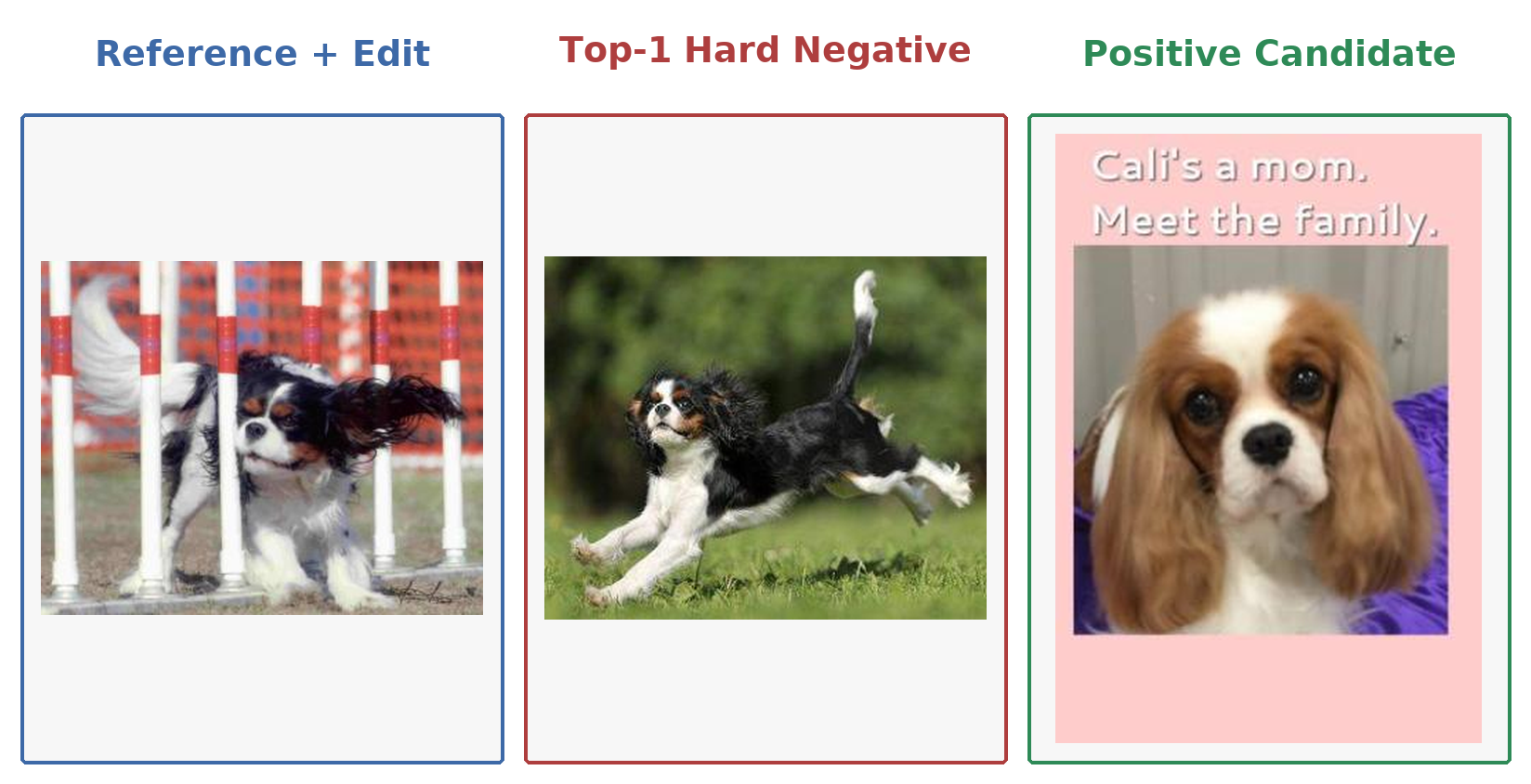}}{}
\end{minipage}\hfill
\begin{minipage}[t]{0.55\textwidth}
\vspace{0pt}
\casecotevidencehead
\scriptsize\raggedright
\fcolorbox{red!60!black}{red!3}{\begin{minipage}{0.965\linewidth}
\textbf{Embedding hard negative: rank 1 $\rightarrow$ 3.}
\textbf{Query intent.} Find an everyday image of a dog that is the same breed as the one in the query image, specifically focusing on its head.
\textbf{Key target observations.}
\cotbullet{The target image shows a dog in mid-motion, running on grass.}
\cotbullet{The image is a dynamic action shot, not a static head shot.}
\cotbullet{The dog's facial features are not clearly visible due to motion blur.}
\end{minipage}}
\par\vspace{3pt}
\fcolorbox{green!55!black}{green!3}{\begin{minipage}{0.965\linewidth}
\textbf{Positive candidate: rank 5 $\rightarrow$ 1.}
\textbf{Query intent.} Find an everyday image of the same breed dog, specifically a Cavalier King Charles Spaniel, focusing on its head.
\textbf{Key target observations.}
\cotbullet{The image shows a Cavalier King Charles Spaniel with long, floppy ears.}
\cotbullet{The dog's head is centered in the frame, showing its facial features.}
\cotbullet{The image is a close-up portrait focusing on the dog's head and upper body.}
\end{minipage}}
\end{minipage}
\end{minipage}

\par\vspace{5pt}\hrule\vspace{5pt}

\begin{minipage}{\textwidth}
\small
\textbf{Case 2: Attribute-sensitive composed retrieval.}
\textbf{Query:} Find an everyday image showing two animals that are of a different species from the first one, with no human appearing.
\par\vspace{3pt}
\begin{minipage}[t]{0.43\textwidth}
\vspace{0pt}
\centering
\casevisualhead
\IfFileExists{supplementary/figures/g-cirr-correction.png}{\includegraphics[width=\linewidth,trim=0 0 0 122,clip]{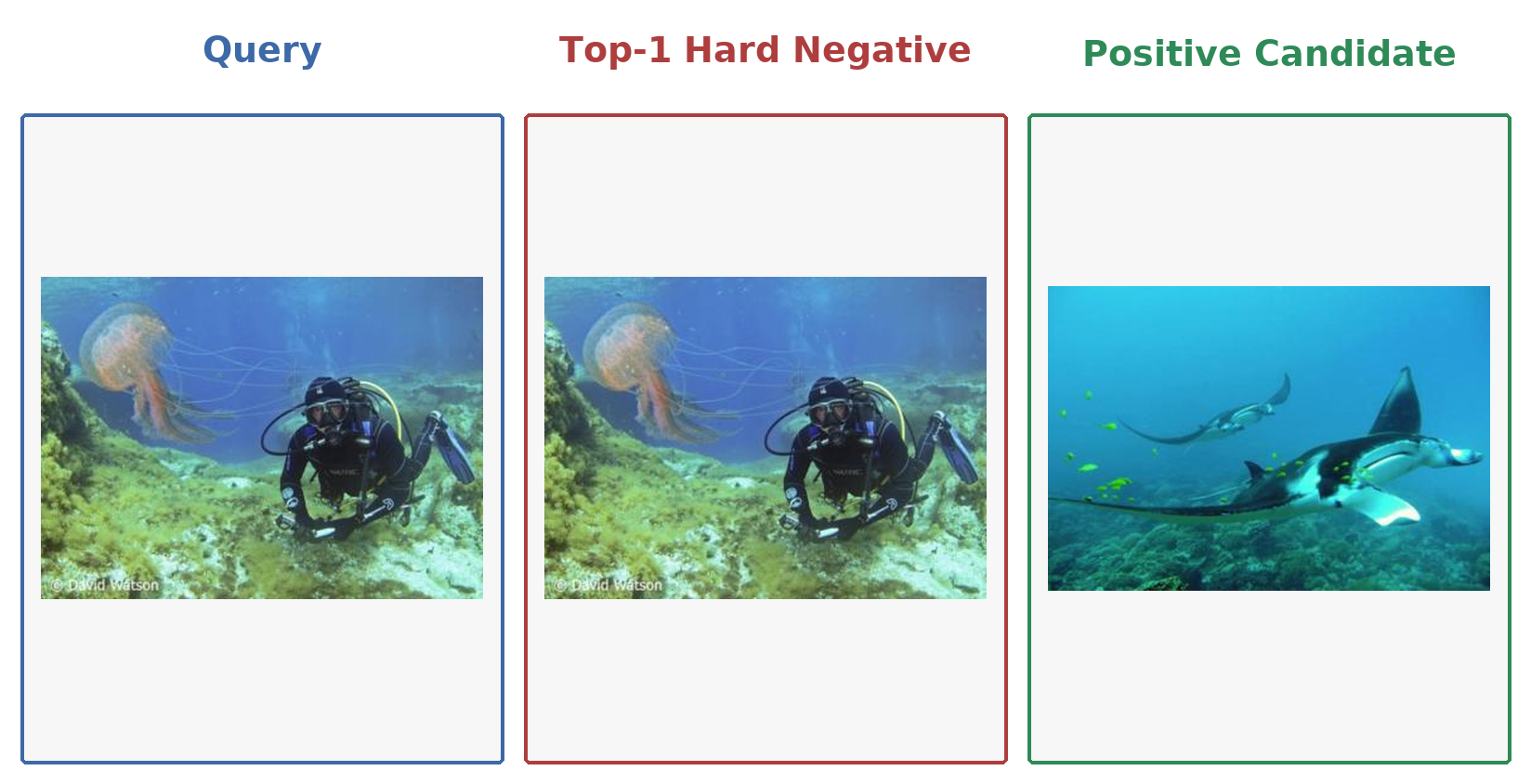}}{}
\end{minipage}\hfill
\begin{minipage}[t]{0.55\textwidth}
\vspace{0pt}
\casecotevidencehead
\scriptsize\raggedright
\fcolorbox{red!60!black}{red!3}{\begin{minipage}{0.965\linewidth}
\textbf{Embedding hard negative: rank 1 $\rightarrow$ 2.}
\textbf{Query intent.} Find an everyday image showing two animals of a different species from the first one, with no human appearing.
\textbf{Key target observations.}
\cotbullet{The image contains one large, orange, jellyfish-like creature.}
\cotbullet{A human is visible underwater, wearing a scuba suit and helmet.}
\cotbullet{The background consists of underwater coral and rocks.}
\end{minipage}}
\par\vspace{3pt}
\fcolorbox{green!55!black}{green!3}{\begin{minipage}{0.965\linewidth}
\textbf{Positive candidate: rank 4 $\rightarrow$ 1.}
\textbf{Query intent.} Find an everyday image showing two animals that are of a different species from the first one, with no human appearing.
\textbf{Key target observations.}
\cotbullet{The image contains two large, black, ray-like marine animals swimming together.}
\cotbullet{There are no human figures visible in the image.}
\cotbullet{The animals are not of the same species as the first one in the query image.}
\end{minipage}}
\end{minipage}
\end{minipage}

\par\vspace{5pt}\hrule\vspace{5pt}

\begin{minipage}{\textwidth}
\small
\textbf{Case 3: Spatial-relation ranking regression.}
\textbf{Query:} Identify the motorcycle behind the red motorcycle on the right.
The green outline marks the requested background motorcycle and the red outline marks the foreground distractor for visualization only.
\par\vspace{3pt}
\begin{minipage}[t]{0.43\textwidth}
\vspace{0pt}
\centering
\casevisualhead
\IfFileExists{supplementary/figures/g-refcoco-spatial-regression.png}{\includegraphics[width=\linewidth,trim=0 0 0 122,clip]{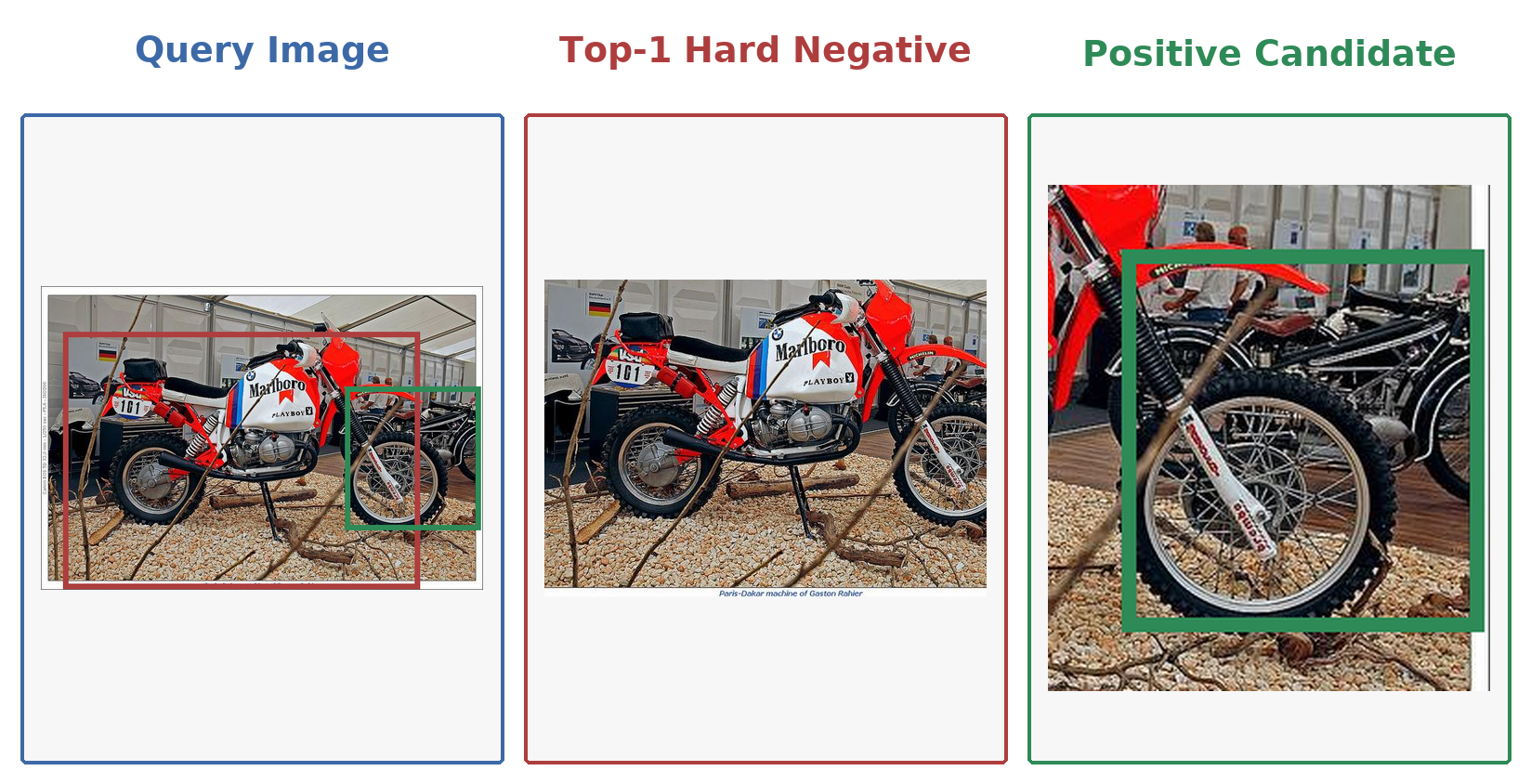}}{}
\end{minipage}\hfill
\begin{minipage}[t]{0.55\textwidth}
\vspace{0pt}
\casecotevidencehead
\scriptsize\raggedright
\fcolorbox{red!60!black}{red!3}{\begin{minipage}{0.965\linewidth}
\textbf{Pair-aware top-1 hard negative: rank 2 $\rightarrow$ 1.}
\textbf{Query intent.} Select the portion of the image that shows the motorcycle closest to the camera, based on its position and visual prominence in the scene.
\textbf{Key target observations.}
\cotbullet{The motorcycle is the central subject, positioned prominently in the foreground.}
\cotbullet{Its front wheel and handlebars are clearly in focus, indicating that it is closest to the camera.}
\cotbullet{It is stationary and appears to be the primary subject of the image.}
\end{minipage}}
\par\vspace{3pt}
\fcolorbox{green!55!black}{green!3}{\begin{minipage}{0.965\linewidth}
\textbf{Positive candidate: rank 1 $\rightarrow$ 2.}
\textbf{Query intent.} Identify the portion of the image corresponding to the ``cycle behind red cycle on right,'' focusing on the motorcycle visible in the background behind the red motorcycle in the foreground.
\textbf{Key target observations.}
\cotbullet{The target refers to a cycle behind the red cycle on the right, indicating a motorcycle behind and to the right of the foreground motorcycle.}
\cotbullet{The phrase ``on right'' matches the orientation of the red motorcycle in the image.}
\cotbullet{No specific brand or model is required by the query.}
\end{minipage}}
\end{minipage}
\end{minipage}

\caption{Two successful corrections and one failure case for pair-aware ranking.
Each image shows the query, competing candidate and positive candidate from left to right; the red and green boxes show their selected CoT evidence.}
\label{fig:qualitative_success_cases}
\end{figure*}

\textbf{Qualitative retrieval corrections.}
Figure~\ref{fig:qualitative_success_cases} shows that pair-aware ranking corrects different forms of embedding confusion.
Case~1 preserves fine-grained breed identity while enforcing the requested head-focused composition, while Case~2 jointly enforces the requested animal count, species change and absence of humans.
The generated evidence explicitly exposes why the embedding top result is insufficient and why the promoted candidate satisfies the query.
Case~3 exposes a complementary failure mode: ranking replaces the requested behind/on-the-right relation with foreground prominence and incorrectly promotes the motorcycle closest to the camera.
This failure delineates a remaining sensitivity to visual salience in spatial-relation queries.

\subsection{Accuracy--Efficiency and System Implications}

\begin{table}[t]
\centering
\scriptsize
\setlength{\tabcolsep}{4pt}
\begin{tabular*}{\columnwidth}
{@{\extracolsep{\fill}}lccccc@{}}
\toprule
\multirow[c]{2}{*}{Model}
& \multirow[c]{2}{*}{$K$}
& \multirow[c]{2}{*}{Score $\uparrow$}
& Reasoning Tokens
& Query Latency
& Indexing Time \\
& & & (/query) & (s/query) $\downarrow$ & (s/candidate) $\downarrow$ \\
\midrule
UME-R1
& -- & 60.1 & 352 & 9.963 & 11.755 \\
PLUME
& -- & 61.6 & 8 & 0.329 & 0.366 \\
\textsc{UMER-E}
& -- & 63.1 & 0 & 0.084 & 0.118 \\
\midrule
\multirow[c]{4}{*}{\textsc{UMER-R}}
& 3 & 64.8 & 410 & 13.566 & \multirow[c]{4}{*}{$0.$} \\
& 5 & 65.5 & 665 & 19.880 & \\
& 10 & 66.0 & 1305 & 42.764 & \\
& 20 & 66.1 & 2508 & 81.726 & \\
\bottomrule
\end{tabular*}
\caption{
Accuracy--efficiency on MMEB-V2 using one NVIDIA A100 GPU
(batch size=1). \textsc{UMER-R} reuses the
\textsc{UMER-E} index; 0. denotes no extra indexing cost.
}
\label{tab:baseline_efficiency}
\end{table}

\textbf{Embedding efficiency.}
Table~\ref{tab:baseline_efficiency} shows that \textsc{UMER-E} avoids the
sequential reasoning required before embedding extraction by UME-R1 and PLUME.
Among the three models in this table, it achieves the highest embedding-only score of 63.1 with zero reasoning tokens,
yielding query/indexing speedups of $118.6\times$/$99.6\times$ over UME-R1 and
$3.9\times$/$3.1\times$ over PLUME.
Since indexing cost scales with corpus size, this advantage is particularly
important for large-scale retrieval.

\textbf{Budget-adjustable inference.}
\textsc{UMER-R} reuses the \textsc{UMER-E} index and spends additional
computation only on online top-$K$ reranking.
As $K$ increases from 3 to 20, tokens and latency grow nearly linearly, whereas
the score saturates from 64.8 to 66.1.
The default $K=5$ reaches 65.5, capturing 80\% of the maximum observed gain
over \textsc{UMER-E} at only 24\% of the $K=20$ latency, defining an explicit
accuracy--latency operating point.

\begin{table}[!b]
\centering
\scriptsize
\setlength{\tabcolsep}{3pt}
\begin{tabular*}{\columnwidth}{@{\extracolsep{\fill}}rrrr@{}}
\toprule
\multirow{2}{*}{Query tokens} & \multicolumn{2}{c}{Rank prefill time (ms) $\downarrow$} & \multirow{2}{*}{Speedup $\uparrow$} \\
\cmidrule(lr){2-3}
& Rebuild Query KV & Reuse retrieval KV & \\
\midrule
256 & 55.39 & 28.06 & $1.97\times$ \\
512 & 56.29 & 28.28 & $1.99\times$ \\
1,024 & 55.16 & 28.17 & $1.96\times$ \\
2,048 & 69.82 & 28.96 & $2.41\times$ \\
4,096 & 110.12 & 29.18 & $3.77\times$ \\
8,192 & 201.61 & 29.08 & $\textbf{6.93}\times$ \\
\bottomrule
\end{tabular*}
\caption{Effect of retrieval KV reuse on ranking-stage prefill time.
Synthetic text-only inputs on one A100, batch size 1.}
\label{tab:query_kv_prefill}
\end{table}

\textbf{Query-KV reuse across retrieval and ranking.}
The shared backbone allows ranking to reuse the query-prefix KV states produced during embedding extraction, avoiding another query prefill.
Table~\ref{tab:query_kv_prefill} reports a synthetic text-only A100 microbenchmark (batch size 1, $K=1$) over 256--8,192-token query prefixes. Each prefix length uses three warmup runs and 20 paired repetitions, with the execution order alternated between the rebuild and reuse settings.
We measure ranking-stage prefill time as synchronized Query-KV setup plus candidate and prompt prefill, including cache branching. 
Reuse avoids repeated query prefill while preserving computation for the new candidate and ranking prompt, keeping the measured cost near 28--29\,ms.
Rebuilding grows from about 55\,ms for short prefixes to 202\,ms at 8,192 tokens. At the longest prefix, reuse saves about 173\,ms and accelerates ranking-stage prefill by $6.93\times$. With fixed 128-token decoding, the corresponding full-ranking speedup is $1.05\times$.




\section{Conclusion}

UMER unifies independently indexable multimodal embeddings and Pair-Aware Discriminative Reasoning within a single MLLM. Pair-Aware CoT provides evidence-bearing supervision, direct ranking strengthens fine-grained relevance judgment, and CMD transfers complementary strengths between embedding and ranking. Experiments across model scales demonstrate consistent improvements over embedding-only retrieval and matched-backbone baselines, while mechanistic analyses reveal improved hard-negative separation and targeted cross-capability transfer. Optional top-$K$ reasoning and query-prefix KV reuse further provide a controllable accuracy--latency tradeoff. Overall, UMER offers a unified path from efficient corpus-scale retrieval to evidence-grounded candidate refinement under different computational budgets. This flexibility supports diverse retrieval needs.

\clearpage

\IfFileExists{aaai2027.bib}{%
  \bibliographystyle{ACM-Reference-Format}
  \bibliography{aaai2027}
}{}

\appendix
\section*{Supplementary Material}
\IfFileExists{supplementary/appendix_for_main.tex}{%
\section{Implementation and Evaluation Details}
Table~\ref{tab:supp-implementation} consolidates the complete implementation and evaluation settings for UMER. The training mixture spans image, video, and visual-document sources, with curated positives replacing raw targets when they are available. Each training instance is paired with a retrieved hard negative for contrastive and pairwise supervision. The pair-aware CoT targets in this mixture are generated and audited offline before training, as detailed below.
\begin{table}[H]
\centering\scriptsize
\setlength{\tabcolsep}{2pt}
\renewcommand{\arraystretch}{0.96}
\begin{tabularx}{\columnwidth}{@{}p{0.36\columnwidth}X@{}}
\toprule
Setting & Value \\
\midrule
\multicolumn{2}{@{}l}{\textit{Model and optimization}} \\
Initialization & Qwen2-VL-2B-Instruct. \\
Embedding tokens & Four learnable tokens per input ($M=4$). \\
Embedding temperature & 0.02. \\
Objective weights & $\lambda_{\rm cot}=0.2$, $\lambda_{\rm bce}=0.1$, $\lambda_{\rm margin}=0.2$, and $\lambda_{\rm cmd}=0.2$. \\
Optimizer and schedule & AdamW; per-device batch size 64; one accumulation step; linear schedule with peak rate $5\times10^{-5}$, 100 warmup steps and 10,000 maximum steps. \\
Adaptation & LoRA with rank 16, scaling 64 and dropout 0.1 applied to attention/MLP projections and the language-model head; visual encoder frozen. \\
Precision and preprocessing & BF16 with FlashAttention-2; maximum image pixels 2,359,296. \\
\midrule
\multicolumn{2}{@{}l}{\textit{Inference}} \\
Candidate selection & Retrieve with the embedding branch and rerank the top $K=5$ candidates. \\
Ranking generation & At most 256 newly generated tokens per candidate. \\
UMER-R score combination & Separately z-score normalize embedding and ranking scores over the five candidates. \\
Ranking-score weight & $\alpha=2.0$. \\
No-CoT ablation & Keep the same positive--hard-negative pairs and BCE/pairwise ranking losses, disable CoT sidecars and CoT conditioning, append $\mathtt{[RANK]}$ directly after the pair prompt, and score each pair in one forward pass; evaluation uses no autoregressive decoding. \\
\midrule
\multicolumn{2}{@{}l}{\textit{Evaluation and reporting}} \\
Benchmark & MMEB-V2: 78 datasets across image, video and visual-document modalities. \\
Benchmark versions & Corrected ViDoSeek-page and MMLongBench-page versions. \\
Metrics & Hit@1 for image and video tasks; NDCG@5 for visual-document tasks. \\
Aggregation & Unweighted macro averages over datasets, including all 78 tasks for \textsc{All}. \\
\bottomrule
\end{tabularx}
\caption{Implementation and evaluation settings for UMER.}
\label{tab:supp-implementation}
\end{table}

\section{Pair-Aware CoT Data Construction}
We construct pair-aware rationales offline in three stages. The pipeline starts with a source-specific query--positive pair, then obtains a semantically close but non-relevant candidate, and finally creates a separate rationale for each of the positive and negative pairs. This makes the supervision suitable for learning which details distinguish the labeled target from a plausible distractor, rather than simply describing a query in isolation.

\textbf{Stage 1: hard-negative mining.} We embed queries and source-local galleries with a frozen Qwen3-VL-Embedding-8B model. For every query, the miner masks all annotated positives and searches the remaining candidates by cosine similarity. To avoid treating likely missing positives as negatives, it discards any candidate whose similarity is at least $0.75$ times the similarity of the annotated positive, and ranks only the remaining candidates. We retain a pool of at most three candidates. For retrieval sources that provide an independently reviewed candidate list, the reviewed non-relevant candidate is preferred, with the embedding-mined pool used as a fallback. The final training configuration takes one offline hard negative per query: the first candidate for the embedding branch and the first candidate with an available CoT record for pairwise reranking.

\textbf{Stage 2: structured CoT generation.} Qwen3.5-9B independently annotates every positive and selected hard-negative pair. It receives the query side and the target side, including any associated image or video, but not the pair label. The output is constrained to two fields: \emph{query intent}, a concrete statement of the query-specific retrieval need, and \emph{target observations}, a list of target-side facts relevant to that need. The prompt explicitly asks for salient overlaps as well as missing or conflicting details, while forbidding a final match/non-match decision. We sample three candidate rationales per pair for image and retrieval sources and one for video and visual-document sources, using temperature $0.7$ and top-$p$ $0.9$. The generation limit is 384 tokens for image/retrieval data and 192 tokens for video/document data. Across all sources, this stage produces 15,539,327 pair-level rationale candidates. During training, a canonical CoT record is retrieved by its normalized query--target pair key; a reranking example is formed only when both its positive and selected negative have such a record.
\noindent
\begin{supppromptbox}[left=5pt,right=5pt,top=4pt,bottom=4pt]
\small
\textbf{Stage 2: image/retrieval CoT generation.}\\[-1pt]
\emph{You will see a retrieval query and a target. Produce objective notes only.}\\[-1pt]
Return valid JSON with two keys: \texttt{query\_intent}, which contains one concrete sentence stating the query's retrieval need and constraints; and \texttt{target\_observations}, which contains three to eight short target-side facts relevant to those constraints. Include relevant entities, attributes, actions, relations, or text cues. Use neutral, non-generic wording and do not state a final match/non-match decision.
\end{supppromptbox}

\textbf{Stage 3: text-only evidence verification.} A Qwen3.5-0.8B auditor sees only the two generated fields---not the original query, target, media, source dataset, or relevance label---and returns \texttt{match}, \texttt{not\_match}, or \texttt{unknown}. A strict audit passes a trace only when its prediction agrees with the pair label, the schema is valid, and no final decision leaks into the rationale. Of the 15,539,327 generated pair records, 8,447,838 pass verification, corresponding to an overall retention rate of 54.4\%. The four sidecars selected for the final training mixture contain 6,632,206 canonical pair records. These counts refer to pair-level records rather than unique queries or positive--negative triplets; the released configuration uses the first schema-valid image/retrieval trace and one video/document trace per pair.
\noindent
\begin{supppromptbox}[left=5pt,right=5pt,top=4pt,bottom=4pt]
\small
\textbf{Stage 3 verifier prompt.} \emph{Use only the generated notes, not the original query, target, media, labels, priors, or outside facts.} Return \texttt{match} only for concrete support, \texttt{not\_match} only for concrete conflict, and \texttt{unknown} otherwise. An asserted decision without observations is insufficient.
\end{supppromptbox}

\section{Limitations and Scope}
UMER's pair-aware supervision depends on hard negatives surfaced by a frozen miner, and its Qwen3.5-0.8B text-only filter measures whether the generated evidence supports the pair label rather than whether every observation is grounded in the source media. This may underrepresent confusions outside the miner's neighborhood or faithful visual evidence that is difficult to verbalize. CMD also selects the teacher at the meta-task level, leaving query-adaptive transfer for future work.

\FloatBarrier

\onecolumn
\section{Complete MMEB-V2 Results}
\subsection{Qwen2-VL-2B Results}
{\raggedright\noindent Table~\ref{tab:supp-full-results} reports every MMEB-V2 task. \textsc{UMER-R} attains the best overall, image, video, and visual-document averages (65.5, 70.4, 45.0, and 73.6), surpassing RIME by 1.4 points overall; detailed task-level outcomes are listed below.\par}
\begingroup
\centering
{\fontsize{5.4}{5.5}\selectfont
\setlength{\tabcolsep}{1.3pt}
\renewcommand{\arraystretch}{0.87}
\begin{tabular*}{\textwidth}{@{\extracolsep{\fill}}l*{10}{r}@{}}
\toprule
Task & GME & VLM2Vec & VLM2Vec-V2 & DUME & BToks & UME-R1 & PLUME & RIME & \textsc{UMER-E} & \textsc{UMER-R} \\
\midrule
\textit{Avg - All (78 tasks)} & 54.1 & 47.0 & 58.0 & 52.7 & 59.0 & 60.1 & 61.6 & \underline{64.1} & 63.1 & \textbf{65.5} \\
\textit{Avg - Image (36 tasks)} & 51.9 & 59.7 & 64.9 & 62.5 & 66.0 & 66.6 & 66.3 & \underline{69.1} & 68.0 & \textbf{70.4} \\
\textit{Avg - Video (18 tasks)} & 33.9 & 29.0 & 34.9 & 33.2 & 39.9 & 42.2 & \underline{44.1} & 43.7 & 41.1 & \textbf{45.0} \\
\textit{Avg - VisDoc (24 tasks)} & \underline{72.7} & 41.6 & 65.4 & 52.8 & 62.7 & 63.9 & 67.5 & 71.4 & 72.2 & \textbf{73.6} \\
\textit{I-CLS (10)} & 54.4 & 58.7 & 62.9 & 59.3 & 64.3 & 64.8 & \underline{66.5} & \textbf{67.9} & 64.4 & 66.1 \\
\textit{I-QA (10)} & 29.9 & 49.3 & 56.3 & 55.0 & 59.8 & 62.8 & 59.2 & 64.4 & \underline{64.9} & \textbf{68.9} \\
\textit{I-RET (12)} & 66.9 & 65.0 & 69.5 & 66.3 & 68.8 & 67.6 & 67.6 & 69.8 & \underline{70.2} & \textbf{71.7} \\
\textit{I-GD (4)} & 55.5 & 72.9 & 77.3 & 78.0 & 77.4 & 77.2 & 79.7 & \textbf{82.1} & 78.4 & \underline{80.8} \\
\textit{V-CLS (5)} & 34.9 & 33.4 & 39.3 & 37.7 & 43.7 & 44.3 & 45.0 & \underline{48.0} & 43.0 & \textbf{50.6} \\
\textit{V-QA (5)} & 42.0 & 30.5 & 34.3 & 46.6 & 47.0 & 51.2 & \underline{52.3} & 52.1 & 50.6 & \textbf{52.8} \\
\textit{V-RET (5)} & 25.6 & 20.6 & 28.8 & 17.1 & 33.0 & 32.9 & 33.5 & 33.6 & \underline{34.7} & \textbf{37.7} \\
\textit{V-MR (3)} & 32.4 & 33.0 & 38.5 & 30.0 & 33.6 & \underline{39.7} & \textbf{46.7} & 39.2 & 32.8 & 35.0 \\
\textit{VD-ViDoRe-V1 (10)} & \textbf{86.1} & 49.8 & 75.5 & 67.6 & 71.1 & 72.4 & 72.1 & 76.4 & 76.1 & \underline{78.0} \\
\textit{VD-ViDoRe-V2 (4)} & \textbf{54.0} & 13.5 & 44.9 & 43.3 & 38.6 & 46.2 & 49.8 & \underline{51.4} & 50.1 & 50.5 \\
\textit{VD-VisRAG (6)} & 82.5 & 51.8 & 79.4 & 47.1 & 81.3 & 79.2 & 78.1 & 81.7 & \underline{82.8} & \textbf{85.0} \\
\textit{VD-OOD (4)} & 43.1 & 33.5 & 39.4 & 33.8 & 38.1 & 37.2 & 57.4 & 63.9 & \underline{68.3} & \textbf{68.8} \\
\midrule
ImageNet-1K & 58.3 & 77.5 & \underline{80.8} & 74.6 & 80.5 & 75.3 & 74.1 & \textbf{81.2} & 80.4 & 80.4 \\
N24News & 50.1 & 73.7 & 72.9 & 69.7 & 74.0 & \textbf{81.1} & \textbf{81.1} & \underline{80.0} & 79.2 & 76.7 \\
HatefulMemes & 52.5 & 58.3 & 56.3 & 65.3 & 62.5 & \underline{75.2} & \textbf{75.5} & 68.4 & 53.3 & 58.9 \\
VOC2007 & 75.9 & 74.3 & 85.0 & 68.9 & 85.7 & 80.0 & \underline{86.1} & \textbf{90.4} & 74.9 & 73.8 \\
SUN397 & 67.3 & 73.8 & 71.0 & 71.4 & 74.9 & \underline{79.4} & 76.9 & \textbf{80.1} & 76.5 & 79.0 \\
Place365 & 35.8 & 35.3 & 35.9 & 41.0 & 38.7 & 42.6 & 42.4 & \textbf{45.3} & 42.4 & \underline{43.4} \\
ImageNet-A & 28.8 & 50.9 & 47.4 & 41.3 & 46.9 & 50.4 & 50.8 & \underline{52.1} & 49.7 & \textbf{56.0} \\
ImageNet-R & 78.6 & 84.7 & 89.3 & \underline{90.7} & 85.6 & 88.7 & 87.5 & 89.9 & 89.4 & \textbf{90.9} \\
ObjectNet & 70.6 & 37.1 & 65.2 & 46.2 & 68.5 & 52.0 & 61.5 & 66.1 & \underline{72.0} & \textbf{76.1} \\
Country211 & \textbf{26.5} & 21.5 & 25.2 & 23.9 & 25.2 & 23.4 & 25.0 & 25.5 & \underline{26.2} & \underline{26.2} \\
OK-VQA & 29.9 & 48.5 & 51.5 & 56.8 & 61.8 & 62.4 & 60.5 & \textbf{65.8} & 58.7 & \underline{63.0} \\
A-OKVQA & 18.6 & 39.5 & 43.6 & 46.9 & 48.6 & 51.1 & 49.9 & \textbf{56.4} & \underline{53.0} & \textbf{56.4} \\
DocVQA & 29.8 & 82.5 & 90.1 & 86.0 & 91.9 & 92.2 & 89.9 & \underline{93.4} & 92.9 & \textbf{94.4} \\
InfographicsVQA & 11.6 & 47.7 & 58.8 & 59.2 & 61.6 & \underline{67.7} & 59.6 & 63.6 & 67.2 & \textbf{70.0} \\
ChartQA & 13.4 & 42.3 & 47.4 & 39.1 & 51.2 & \textbf{64.9} & 49.8 & \underline{61.7} & 54.3 & 57.6 \\
Visual7W & 16.2 & 51.2 & 52.9 & 46.9 & 49.2 & 54.1 & 47.6 & 54.8 & \underline{64.5} & \textbf{66.0} \\
ScienceQA & 27.3 & 30.7 & 38.2 & 38.7 & 40.1 & 42.7 & 42.9 & \underline{45.7} & 44.8 & \textbf{49.8} \\
VizWiz & 37.0 & 38.6 & 43.3 & 42.0 & 49.8 & 46.8 & 46.5 & 47.7 & \underline{52.5} & \textbf{57.4} \\
GQA & 75.1 & 48.3 & 64.9 & 60.2 & 63.9 & 67.3 & 69.1 & 73.1 & \underline{77.5} & \textbf{86.5} \\
TextVQA & 39.7 & 63.3 & 72.2 & 73.9 & 79.5 & 78.6 & 78.9 & 81.3 & \underline{83.6} & \textbf{87.5} \\
VisDial & 48.1 & 74.3 & \underline{82.7} & 75.9 & 78.4 & 76.6 & 72.6 & 80.7 & 82.4 & \textbf{85.9} \\
CIRR & 44.2 & 46.8 & \underline{57.5} & 52.0 & 54.0 & 53.7 & 54.6 & 57.0 & 52.9 & \textbf{57.9} \\
VisualNews\_t2i & 74.7 & 73.1 & 74.5 & 71.2 & 72.7 & 71.7 & 71.3 & 72.4 & \underline{76.0} & \textbf{76.6} \\
VisualNews\_i2t & 78.3 & 73.7 & 78.2 & 72.5 & 75.8 & 74.2 & 72.7 & 78.2 & \underline{79.1} & \textbf{82.1} \\
MSCOCO\_t2i & 68.1 & 73.4 & 75.3 & 74.5 & 73.9 & 75.1 & 74.1 & \underline{76.1} & 76.0 & \textbf{77.3} \\
MSCOCO\_i2t & 63.1 & 68.5 & 71.4 & 68.3 & 69.1 & 68.9 & 69.8 & 70.0 & \underline{73.0} & \textbf{75.0} \\
NIGHTS & 67.0 & 66.3 & \textbf{68.6} & 67.5 & \underline{68.1} & 67.2 & 68.0 & 67.9 & 64.0 & 63.7 \\
WebQA & 88.8 & 85.9 & 90.6 & 90.2 & 90.5 & 90.0 & 89.1 & \underline{90.7} & 90.1 & \textbf{92.1} \\
FashionIQ & \textbf{32.9} & 14.0 & 19.5 & 11.5 & 19.1 & 17.1 & \underline{20.3} & 19.8 & 20.0 & 17.2 \\
Wiki-SS-NQ & \textbf{73.9} & 54.2 & 66.9 & 60.0 & 70.5 & 62.0 & 68.6 & 68.9 & 70.9 & \underline{71.8} \\
OVEN & \textbf{72.3} & 68.3 & 64.3 & 65.2 & 67.8 & 66.9 & 68.4 & 67.5 & \underline{69.1} & 68.1 \\
EDIS & \underline{91.8} & 81.2 & 84.1 & 86.5 & 85.4 & 88.0 & 81.8 & 88.9 & 88.9 & \textbf{92.6} \\
MSCOCO & 28.6 & 66.5 & 67.1 & 68.1 & 66.4 & \textbf{69.5} & 66.9 & 69.0 & 67.7 & \underline{69.1} \\
RefCOCO & 55.9 & 80.9 & 87.1 & 85.1 & 87.2 & 83.3 & 86.5 & \underline{88.9} & 87.7 & \textbf{90.3} \\
RefCOCO-Matching & 73.3 & 75.7 & 85.8 & \underline{89.3} & 86.4 & 84.4 & 88.4 & \textbf{90.1} & 85.7 & 81.1 \\
Visual7W-Pointing & 64.1 & 68.3 & 69.2 & 69.5 & 69.4 & 71.5 & 74.9 & \underline{80.5} & 72.6 & \textbf{82.5} \\
\midrule
K700 & 35.2 & 31.4 & 38.0 & 22.7 & 43.1 & 35.8 & 42.2 & \underline{47.7} & 45.6 & \textbf{50.6} \\
SmthSmthV2 & 29.9 & 30.9 & 42.8 & 37.7 & 41.0 & 44.1 & 44.8 & \underline{48.5} & 43.7 & \textbf{54.7} \\
HMDB51 & 43.4 & 33.8 & 40.9 & 53.4 & 47.1 & \underline{54.4} & 51.2 & \textbf{56.7} & 42.3 & 53.8 \\
UCF101 & 52.4 & 57.5 & 60.0 & 55.7 & \textbf{69.3} & 67.2 & 66.5 & 66.4 & 62.9 & \underline{68.1} \\
Breakfast & 13.6 & 13.4 & 14.8 & 18.9 & 18.0 & 20.1 & 20.1 & \underline{20.6} & 20.3 & \textbf{25.9} \\
MVBench & 37.5 & 30.5 & 33.7 & 48.8 & 45.4 & \underline{49.9} & 47.4 & 49.1 & 48.6 & \textbf{51.7} \\
Video-MME & 34.3 & 26.9 & 30.7 & 39.2 & 39.9 & \textbf{41.7} & 40.0 & \underline{41.6} & 39.3 & 40.3 \\
NExTQA & 39.5 & 20.3 & 20.9 & 55.2 & 47.9 & \textbf{59.9} & 57.3 & \underline{58.9} & 51.4 & 58.3 \\
EgoSchema & 40.8 & 25.4 & 34.0 & 23.2 & 37.0 & 45.4 & \underline{47.8} & 40.4 & 42.6 & \textbf{49.4} \\
ActivityNetQA & 58.0 & 49.6 & 52.3 & 66.7 & 64.8 & 57.8 & 69.2 & \underline{70.5} & \textbf{70.9} & 64.3 \\
DiDeMo & 22.0 & 19.4 & 30.4 & 16.9 & 33.0 & 32.4 & 32.7 & \underline{33.8} & 33.2 & \textbf{34.9} \\
MSR-VTT & 27.3 & 25.2 & 28.3 & 16.2 & 33.8 & 34.3 & 36.2 & 36.4 & \underline{37.8} & \textbf{41.9} \\
MSVD & 47.6 & 38.2 & 48.1 & 34.9 & 56.0 & 55.4 & 56.1 & 56.6 & \underline{59.0} & \textbf{61.3} \\
VATEX & 23.0 & 16.2 & 26.5 & 11.1 & 27.6 & \underline{29.9} & 28.2 & 28.1 & 29.5 & \textbf{32.5} \\
YouCook2 & 7.9 & 4.1 & 10.6 & 0.1 & \underline{14.5} & 12.7 & \underline{14.5} & 13.3 & 14.2 & \textbf{17.8} \\
QVHighlight & 43.6 & 44.2 & 49.4 & 40.3 & 42.2 & \textbf{57.5} & \underline{57.1} & 55.1 & 40.9 & 46.3 \\
Charades-STA & 14.9 & 13.6 & \underline{20.2} & 16.1 & 18.0 & \textbf{20.4} & 19.4 & 19.4 & 15.5 & 15.3 \\
MomentSeeker & 34.8 & 34.4 & 40.8 & 33.7 & 40.4 & 41.2 & \textbf{63.5} & 43.2 & 42.0 & \underline{43.4} \\
\midrule
ViDoRe\_arxivqa & \textbf{82.8} & 48.9 & 80.6 & 68.7 & 76.5 & 73.9 & 72.6 & \underline{82.1} & 79.1 & 80.5 \\
ViDoRe\_docvqa & \textbf{53.1} & 27.0 & 44.9 & 33.6 & 37.2 & 37.9 & 36.2 & 45.4 & 43.5 & \underline{46.0} \\
ViDoRe\_infovqa & \textbf{90.2} & 67.2 & 83.7 & 74.5 & 80.9 & 76.2 & 79.0 & 81.2 & 82.8 & \underline{84.1} \\
ViDoRe\_tabfquad & \textbf{93.3} & 62.6 & \underline{89.2} & 78.3 & 80.4 & 86.1 & 88.8 & 85.5 & 86.7 & 88.1 \\
ViDoRe\_tatdqa & \textbf{69.9} & 19.8 & 43.8 & 35.3 & 42.0 & 40.6 & 36.6 & 48.9 & 47.2 & \underline{50.0} \\
ViDoRe\_shiftproject & \textbf{89.5} & 41.8 & 60.8 & 61.8 & 62.3 & 66.8 & 64.8 & \underline{69.9} & 69.2 & 69.2 \\
ViDoRe\_syntheticDocQA\_artificial\_intelligence & \textbf{97.5} & 55.0 & 88.5 & 74.3 & 79.9 & 85.9 & 83.8 & 89.8 & 86.6 & \underline{91.3} \\
ViDoRe\_syntheticDocQA\_energy & \textbf{91.9} & 59.1 & 86.5 & 78.4 & 85.3 & 83.3 & 82.6 & 86.2 & 86.5 & \underline{86.9} \\
ViDoRe\_syntheticDocQA\_government\_reports & \textbf{94.6} & 57.1 & 85.0 & 83.0 & 81.0 & 82.6 & 83.2 & 86.4 & 89.1 & \underline{91.7} \\
ViDoRe\_syntheticDocQA\_healthcare\_industry & \textbf{98.7} & 59.6 & 92.2 & 88.2 & 85.3 & 90.8 & 91.1 & 88.3 & 90.7 & \underline{92.3} \\
ViDoRe\_esg\_reports\_human\_labeled\_v2 & \textbf{61.0} & 12.6 & 45.6 & 48.0 & 40.0 & 50.2 & 52.1 & \underline{56.5} & 53.4 & 53.8 \\
ViDoRe\_biomedical\_lectures\_v2\_multilingual & \textbf{54.0} & 7.4 & 44.3 & 39.8 & 39.1 & 46.2 & 48.2 & 47.6 & 47.8 & \underline{50.1} \\
ViDoRe\_economics\_reports\_v2\_multilingual & \textbf{50.2} & 13.9 & 43.0 & 44.1 & 39.7 & 45.7 & \underline{49.6} & 47.0 & 46.8 & 45.8 \\
ViDoRe\_esg\_reports\_v2\_multilingual & 50.7 & 20.1 & 46.6 & 41.1 & 35.7 & 42.7 & 49.0 & \textbf{54.3} & \underline{52.5} & 52.4 \\
VisRAG\_ArxivQA & \textbf{82.0} & 41.8 & 76.9 & 35.8 & 76.9 & 74.3 & 71.6 & 79.4 & 79.5 & \underline{81.2} \\
VisRAG\_ChartQA & 79.9 & 57.9 & 84.4 & 47.2 & 86.2 & 86.0 & 80.8 & 83.3 & \underline{87.8} & \textbf{89.2} \\
VisRAG\_MP-DocVQA & \textbf{84.4} & 43.2 & 71.8 & 35.3 & 78.6 & 75.6 & 74.9 & 82.3 & 81.4 & \underline{84.3} \\
VisRAG\_SlideVQA & \underline{93.4} & 74.0 & 91.5 & 61.3 & 91.8 & 87.1 & 88.9 & 91.0 & 91.5 & \textbf{93.5} \\
VisRAG\_InfoVQA & \textbf{91.4} & 70.7 & 85.7 & 64.7 & 88.7 & 84.4 & 85.7 & 82.4 & 87.4 & \underline{89.9} \\
VisRAG\_PlotQA & 64.1 & 23.4 & 66.1 & 38.5 & 65.6 & 68.0 & 66.2 & \underline{71.6} & 69.5 & \textbf{72.0} \\
ViDoSeek-page & 21.6 & 17.7 & 21.9 & 20.0 & 21.7 & 21.2 & 80.6 & 80.7 & \underline{81.0} & \textbf{82.3} \\
ViDoSeek-doc & 83.6 & 74.3 & 80.2 & 69.5 & 78.3 & 75.9 & 76.9 & 79.7 & \underline{90.7} & \textbf{90.8} \\
MMLongBench-page & 15.8 & 9.6 & 11.9 & 10.4 & 11.0 & 11.9 & 39.9 & \textbf{48.3} & 47.9 & \underline{48.1} \\
MMLongBench-doc & 51.4 & 32.6 & 43.7 & 35.4 & 41.4 & 39.7 & 32.0 & 46.9 & \underline{53.7} & \textbf{53.9} \\
\bottomrule
\end{tabular*}
}
\scriptsize
\captionof{table}{Complete 78-task MMEB-V2 comparison of Qwen2-VL-2B models. Only baselines with published 2B task-level results are included. E denotes embedding-only retrieval; R denotes pair-aware reranking followed by weighted combination of the ranking and embedding scores. Image/video use Hit@1 and visual documents use NDCG@5. Best and second-best values are \textbf{bold} and \underline{underlined}.}
\label{tab:supp-full-results}
\par
\endgroup

\subsection{Qwen2-VL-7B Results}
Table~\ref{tab:supp-full-results-7b} compares Qwen2-VL-7B models with published 78-task breakdowns. With the same $K=5$ score-combination configuration, \textsc{UMER-R} reaches 70.6 overall and 74.3/50.5/80.0 on image/video/visual-document tasks.

\begingroup
\centering
{\fontsize{5.4}{5.5}\selectfont
\setlength{\tabcolsep}{1.9pt}
\renewcommand{\arraystretch}{0.87}
\begin{tabular*}{\textwidth}{@{\extracolsep{\fill}}l*{7}{r}@{}}
\toprule
Task & GME-7B & VLM2Vec-7B & DUME-7B & UME-R1-7B & RIME-7B & \textsc{UMER-E} & \textsc{UMER-R} \\
\midrule
\textit{Avg - All (78 tasks)} & 57.8 & 52.3 & 55.9 & 64.5 & \underline{68.6} & 67.0 & \textbf{70.6} \\
\textit{Avg - Image (36 tasks)} & 56.0 & 65.5 & 66.4 & 71.3 & \underline{73.4} & 71.2 & \textbf{74.3} \\
\textit{Avg - Video (18 tasks)} & 38.4 & 33.7 & 29.4 & 47.5 & \underline{49.4} & 45.2 & \textbf{50.5} \\
\textit{Avg - VisDoc (24 tasks)} & 75.2 & 46.4 & 60.3 & 67.1 & 75.6 & \underline{76.8} & \textbf{80.0} \\
\textit{I-CLS (10)} & 57.7 & 62.7 & 64.2 & 67.1 & \textbf{70.3} & 66.2 & \underline{68.0} \\
\textit{I-QA (10)} & 34.7 & 56.9 & 57.0 & 69.2 & \underline{71.7} & 69.7 & \textbf{73.7} \\
\textit{I-RET (12)} & 71.2 & 69.4 & 70.8 & 71.9 & \underline{73.2} & 72.3 & \textbf{75.6} \\
\textit{I-GD (4)} & 59.3 & 82.2 & 81.8 & 84.9 & \underline{86.3} & 84.6 & \textbf{87.7} \\
\textit{V-CLS (5)} & 37.4 & 39.1 & 32.9 & 48.6 & \underline{52.6} & 48.0 & \textbf{55.2} \\
\textit{V-QA (5)} & 50.4 & 30.0 & 47.4 & \underline{60.7} & \textbf{62.0} & 54.7 & \underline{60.7} \\
\textit{V-RET (5)} & 28.4 & 29.0 & 8.6 & 38.2 & \underline{38.4} & 35.6 & \textbf{40.5} \\
\textit{V-MR (3)} & 37.0 & 38.9 & 28.0 & 39.3 & \underline{41.6} & 40.7 & \textbf{42.5} \\
\textit{VD-ViDoRe-V1 (10)} & \textbf{89.4} & 56.9 & 67.1 & 75.7 & 80.9 & 81.3 & \underline{85.0} \\
\textit{VD-ViDoRe-V2 (4)} & 55.6 & 9.4 & 35.2 & 50.5 & 55.6 & \underline{57.8} & \textbf{62.1} \\
\textit{VD-VisRAG (6)} & 85.0 & 59.1 & 82.6 & 83.7 & \underline{85.8} & 85.1 & \textbf{87.9} \\
\textit{VD-OOD (4)} & 44.4 & 38.1 & 34.9 & 37.6 & 66.9 & \underline{72.5} & \textbf{73.8} \\
\midrule
ImageNet-1K & 64.6 & 80.1 & 76.6 & 80.4 & 80.9 & \underline{82.0} & \textbf{82.8} \\
N24News & 50.5 & 79.7 & 77.2 & \underline{82.3} & \textbf{82.7} & 80.8 & 78.5 \\
HatefulMemes & 53.6 & 69.7 & \textbf{79.6} & \underline{79.0} & 76.2 & 54.1 & 60.2 \\
VOC2007 & 80.3 & 80.7 & 85.5 & \underline{90.8} & \textbf{91.0} & 74.1 & 76.5 \\
SUN397 & 69.5 & 77.4 & 74.6 & \underline{80.3} & \textbf{80.6} & 77.0 & 79.9 \\
Place365 & 39.1 & 37.4 & 41.9 & \textbf{46.8} & \underline{45.5} & 43.8 & 44.1 \\
ImageNet-A & 41.2 & \underline{58.1} & 48.6 & 53.9 & 57.4 & 58.0 & \textbf{60.9} \\
ImageNet-R & 83.9 & 73.9 & 88.8 & 90.1 & 89.9 & \underline{90.8} & \textbf{92.2} \\
ObjectNet & 69.0 & 40.1 & 44.8 & 42.3 & 72.7 & \underline{74.5} & \textbf{78.0} \\
Country211 & 24.8 & \textbf{29.8} & 24.7 & 25.0 & 26.4 & 26.7 & \underline{27.3} \\
OK-VQA & 33.2 & 56.8 & 61.6 & 71.7 & \textbf{74.1} & 67.6 & \underline{71.9} \\
A-OKVQA & 21.0 & 47.3 & 51.4 & 58.7 & \textbf{61.8} & 59.1 & \underline{61.4} \\
DocVQA & 41.4 & 89.7 & 86.3 & 93.8 & 94.4 & \underline{94.7} & \textbf{95.7} \\
InfographicsVQA & 20.3 & 60.0 & 62.3 & \textbf{79.2} & \underline{79.1} & 76.6 & \textbf{79.2} \\
ChartQA & 17.8 & 56.9 & 49.8 & \underline{75.1} & \textbf{77.4} & 64.2 & 64.9 \\
Visual7W & 22.2 & 52.7 & 52.1 & 55.2 & 54.9 & \underline{64.9} & \textbf{69.1} \\
ScienceQA & 28.0 & 38.5 & 45.5 & 53.7 & \textbf{59.0} & 50.7 & \underline{58.9} \\
VizWiz & 39.0 & 39.9 & 44.3 & 51.6 & 55.3 & \underline{56.2} & \textbf{60.9} \\
GQA & \underline{76.9} & 55.1 & 46.9 & 69.3 & 73.6 & 76.1 & \textbf{86.5} \\
TextVQA & 46.8 & 71.6 & 69.9 & 83.5 & \underline{87.0} & \underline{87.0} & \textbf{88.8} \\
VisDial & 60.8 & 81.9 & 75.7 & 80.7 & 82.6 & \underline{83.1} & \textbf{88.4} \\
CIRR & 54.9 & 51.1 & 51.6 & 55.3 & \textbf{60.0} & 54.0 & \underline{59.0} \\
VisualNews\_t2i & 79.7 & \underline{80.5} & 76.9 & 76.8 & 79.8 & 78.8 & \textbf{81.4} \\
VisualNews\_i2t & \underline{83.6} & 81.2 & 82.3 & 82.0 & 83.5 & 82.1 & \textbf{84.5} \\
MSCOCO\_t2i & 71.2 & 77.2 & 77.1 & 78.3 & 77.8 & \underline{78.5} & \textbf{81.6} \\
MSCOCO\_i2t & 57.7 & \underline{73.9} & 71.2 & 71.4 & 72.6 & 73.6 & \textbf{76.2} \\
NIGHTS & 67.6 & 67.6 & \textbf{69.6} & 68.1 & \underline{68.6} & 64.9 & 67.8 \\
WebQA & \textbf{91.4} & 88.3 & 90.3 & \underline{90.9} & 90.7 & 89.8 & \textbf{91.4} \\
FashionIQ & \textbf{37.8} & 17.1 & 20.5 & 23.4 & 26.0 & 27.1 & \underline{29.6} \\
Wiki-SS-NQ & \underline{78.2} & 62.3 & 70.6 & 72.5 & 76.3 & 73.6 & \textbf{79.7} \\
OVEN & \textbf{75.1} & 66.5 & 70.5 & 71.4 & 68.6 & 72.1 & \underline{73.0} \\
EDIS & \textbf{96.0} & 85.7 & 92.8 & 92.0 & 91.6 & 89.9 & \underline{94.5} \\
MSCOCO & 31.4 & \textbf{75.7} & 72.3 & 72.7 & 72.0 & 72.5 & \underline{75.2} \\
RefCOCO & 60.9 & 87.6 & 86.8 & 91.4 & 91.7 & \underline{92.3} & \textbf{94.2} \\
RefCOCO-Matching & 78.4 & 84.6 & 85.1 & 91.1 & \textbf{93.7} & \underline{92.8} & 90.8 \\
Visual7W-Pointing & 66.5 & 81.0 & 83.1 & 84.2 & \underline{87.9} & 80.6 & \textbf{90.4} \\
\midrule
K700 & 39.7 & 35.5 & 27.3 & 42.8 & \underline{55.0} & 51.5 & \textbf{55.8} \\
SmthSmthV2 & 30.6 & 32.1 & 25.1 & 50.4 & \underline{55.1} & 49.7 & \textbf{60.3} \\
HMDB51 & 47.9 & 42.2 & 42.6 & \underline{58.3} & \textbf{58.9} & 50.3 & \textbf{58.9} \\
UCF101 & 54.7 & 61.8 & 48.8 & \underline{70.0} & 68.5 & 64.2 & \textbf{72.7} \\
Breakfast & 14.3 & 23.8 & 20.8 & 21.5 & \underline{25.4} & 24.2 & \textbf{28.4} \\
MVBench & 46.6 & 28.5 & 47.4 & \underline{58.2} & \textbf{59.9} & 52.8 & 57.5 \\
Video-MME & 39.2 & 27.8 & 40.2 & \underline{47.3} & \textbf{49.6} & 44.1 & 46.4 \\
NExTQA & 53.6 & 20.3 & 48.6 & \underline{69.6} & \textbf{69.7} & 55.3 & 69.1 \\
EgoSchema & 46.8 & 21.8 & 50.4 & 52.4 & \underline{55.6} & 50.2 & \textbf{58.0} \\
ActivityNetQA & 65.6 & 51.4 & 50.2 & \textbf{76.0} & \underline{75.1} & 70.9 & 72.3 \\
DiDeMo & 26.4 & 29.3 & 0.1 & \underline{40.0} & 38.4 & 36.2 & \textbf{40.4} \\
MSR-VTT & 31.8 & 34.5 & 0.1 & 38.9 & \underline{41.5} & 36.9 & \textbf{43.7} \\
MSVD & 49.7 & 46.7 & 28.8 & \underline{60.8} & 59.4 & 58.1 & \textbf{63.9} \\
VATEX & 24.9 & 25.5 & 13.8 & 32.6 & \underline{32.7} & 31.2 & \textbf{35.2} \\
YouCook2 & 9.1 & 9.0 & 0.0 & 18.5 & \textbf{19.9} & 15.9 & \underline{19.2} \\
QVHighlight & \textbf{59.5} & \underline{57.7} & 29.4 & 54.9 & 56.9 & 56.1 & 56.6 \\
Charades-STA & 14.0 & 19.8 & 15.8 & \textbf{21.9} & \underline{21.6} & 19.9 & 20.5 \\
MomentSeeker & 37.4 & 39.3 & 38.8 & 41.1 & \underline{46.2} & 45.9 & \textbf{50.4} \\
\midrule
ViDoRe\_arxivqa & \textbf{86.9} & 60.2 & 66.6 & 73.6 & 84.1 & 83.8 & \underline{86.0} \\
ViDoRe\_docvqa & \textbf{57.5} & 34.7 & 35.8 & 41.1 & 46.9 & 48.0 & \underline{51.5} \\
ViDoRe\_infovqa & \textbf{91.6} & 70.4 & 72.8 & 80.8 & 85.4 & 86.8 & \underline{90.3} \\
ViDoRe\_tabfquad & 94.6 & 78.2 & 89.2 & 90.2 & \underline{95.3} & 94.7 & \textbf{96.1} \\
ViDoRe\_tatdqa & \textbf{74.1} & 27.6 & 38.5 & 46.7 & 52.6 & 54.6 & \underline{60.9} \\
ViDoRe\_shiftproject & \textbf{96.8} & 38.6 & 61.9 & 65.0 & 73.7 & 78.4 & \underline{80.9} \\
ViDoRe\_syntheticDocQA\_artificial\_intelligence & \textbf{99.6} & 67.7 & 69.3 & 89.5 & 96.1 & 91.7 & \underline{96.4} \\
ViDoRe\_syntheticDocQA\_energy & \textbf{95.3} & 60.4 & 68.4 & 85.7 & 88.5 & 88.9 & \underline{92.3} \\
ViDoRe\_syntheticDocQA\_government\_reports & \textbf{98.8} & 61.8 & 83.1 & 89.8 & 91.7 & 91.9 & \underline{96.4} \\
ViDoRe\_syntheticDocQA\_healthcare\_industry & \textbf{99.3} & 69.9 & 84.9 & 94.3 & 94.9 & 93.9 & \underline{98.8} \\
ViDoRe\_esg\_reports\_human\_labeled\_v2 & 63.4 & 6.8 & 40.4 & 50.4 & 60.2 & \underline{63.9} & \textbf{69.9} \\
ViDoRe\_biomedical\_lectures\_v2\_multilingual & 49.5 & 5.1 & 37.4 & 50.7 & 51.5 & \underline{56.5} & \textbf{60.4} \\
ViDoRe\_economics\_reports\_v2\_multilingual & 54.2 & 13.9 & 29.6 & \underline{57.8} & \textbf{59.2} & 54.6 & 56.5 \\
ViDoRe\_esg\_reports\_v2\_multilingual & 55.4 & 11.9 & 33.5 & 43.2 & 51.6 & \underline{56.1} & \textbf{61.5} \\
VisRAG\_ArxivQA & \textbf{87.4} & 52.6 & 77.3 & 80.5 & 84.0 & 83.3 & \underline{85.5} \\
VisRAG\_ChartQA & 81.9 & 70.2 & 83.4 & 85.0 & 85.4 & \underline{85.8} & \textbf{89.1} \\
VisRAG\_MP-DocVQA & \textbf{89.2} & 52.8 & 83.8 & 83.4 & 87.4 & 85.0 & \underline{88.5} \\
VisRAG\_SlideVQA & \underline{94.5} & 72.8 & 91.5 & 91.5 & 94.1 & 94.0 & \textbf{95.8} \\
VisRAG\_InfoVQA & \underline{93.5} & 72.0 & 88.2 & 89.2 & 91.8 & 90.9 & \textbf{93.9} \\
VisRAG\_PlotQA & 63.4 & 34.4 & 71.3 & \underline{72.7} & 72.1 & 71.6 & \textbf{74.3} \\
ViDoSeek-page & 23.2 & 22.3 & 20.2 & 21.3 & \underline{85.6} & \underline{85.6} & \textbf{88.9} \\
ViDoSeek-doc & 83.9 & 77.8 & 73.2 & 75.3 & 80.9 & \underline{93.0} & \textbf{93.1} \\
MMLongBench-page & 16.2 & 11.8 & 10.3 & 12.3 & 52.4 & \underline{53.6} & \textbf{55.1} \\
MMLongBench-doc & 54.3 & 40.5 & 36.0 & 41.3 & 48.8 & \underline{57.7} & \textbf{58.2} \\
\bottomrule
\end{tabular*}
}
\scriptsize
\enlargethispage{\baselineskip}
\captionof{table}{Complete 78-task MMEB-V2 comparison of Qwen2-VL-7B models. Only baselines with published 7B task-level results are included. E denotes embedding-only retrieval; R denotes pair-aware reranking followed by weighted combination of the ranking and embedding scores. Image/video use Hit@1 and visual documents use NDCG@5. Best and second-best values are \textbf{bold} and \underline{underlined}.\label{tab:supp-full-results-7b}}
\endgroup

}{}

\end{document}